\documentclass[runningheads]{llncs}
\usepackage{graphicx}

\usepackage{tikz}
\usepackage{comment}
\usepackage{amsmath,amssymb} 
\usepackage{color}

\usepackage{bm}
\usepackage{booktabs}
\usepackage{array}

\usepackage{stmaryrd}
\usepackage{colortbl}
\usepackage{xcolor}
\definecolor{tbgray}{rgb}{0.9,0.97,1.0}
\usepackage{multirow}
\usepackage{wrapfig}
\usepackage{subfigure}
\usepackage{url}

\usepackage{float}
\newfloat{figtab}{htb}{fgtb}
\makeatletter
  \newcommand\figcaption{\def\@captype{figure}\caption}
  \newcommand\tabcaption{\def\@captype{table}\caption}
\makeatother

\usepackage[accsupp]{axessibility}  

\newcommand{\HL}[1]{\textcolor[rgb]{0.8,0.31,0.2235 }{\textbf{#1}}}
\newcommand{\etal}[0]{\textit{et al.}}
\newcommand{\eg}[0]{\textit{e.g.}}
\newcommand{\etc}[0]{\textit{etc}}
\newcommand{\ie}[0]{\textit{i.e.}}

\begin{document}
\pagestyle{headings}
\mainmatter
\def\ECCVSubNumber{2872}  

\title{Prior Knowledge Guided Unsupervised Domain Adaptation} 


\titlerunning{Prior Knowledge Guided Unsupervised Domain Adaptation}
%
\author{Tao Sun\inst{1} \and
Cheng Lu\inst{2} \and
Haibin Ling\inst{1}}
\authorrunning{Sun et al.}
%
\institute{$^1$Stony Brook University \quad
$^2$XPeng Motors\\
\email{\{tao,hling\}@cs.stonybrook.edu, luc@xiaopeng.com}\\}
\maketitle

\begin{abstract}

The waive of labels in the target domain makes Unsupervised Domain Adaptation (UDA) an attractive technique in many real-world applications, though it also brings great challenges as model adaptation becomes harder without labeled target data. In this paper, we address this issue by seeking compensation from target domain prior knowledge, which is often (partially) available in practice, \eg, from human expertise. This leads to a novel yet practical setting where in addition to the training data, some prior knowledge about the target class distribution are available. We term the setting as Knowledge-guided Unsupervised Domain Adaptation (KUDA). In particular, we consider two specific types of prior knowledge about the class distribution in the target domain: \emph{Unary Bound} that describes the lower and upper bounds of individual class probabilities, and \emph{Binary Relationship} that describes the relations between two class probabilities. We propose a general rectification module that uses such prior knowledge to refine model generated pseudo labels. The module is formulated as a Zero-One Programming problem derived from the prior knowledge and a smooth regularizer. It can be easily plugged into self-training based UDA methods, and we combine it with two state-of-the-art methods, SHOT and DINE. Empirical results on four benchmarks confirm that the rectification module clearly improves the quality of pseudo labels, which in turn benefits the self-training stage. With the guidance from prior knowledge, the performances of both methods are substantially boosted. We expect our work to inspire further investigations in integrating prior knowledge in UDA. Code is available at \url{https://github.com/tsun/KUDA}.

\end{abstract}

\section{Introduction}
Deep neural networks have shown significant performance improvement in a variety of vision tasks~\cite{he2016deep,ren2015faster,long2015fully,lecun2015deep}. However, such performance highly relies on massive annotated data, which is often expensive to obtain. Unsupervised Domain Adaptation (UDA) addresses this issue by transferring a predictive model learned from a labeled source domain to an unlabeled target domain~\cite{pan2009survey,wang2018deep,wilson2020survey}. Despite the advancement made in recent years, UDA remains a challenging task due to the absence of labels in the target domain. 
On the other hand, in many real-world applications, prior knowledge about the target domain is often readily available. In particular, some information about class distribution is often available without bothering labeling specific target samples. 
For example, botanists can estimate the proportion of wild species within a reserve using historical information; economists can tell whether vans are more possessed than other vehicles based on the local industrial structure; \etc. Such prior knowledge may provide valuable clues that are complementary to the unlabeled training data, and can be especially beneficial when there exists a large distribution shift between source and target domains. In fact, prior knowledge has been used to compensate the deficiency of labeled data~\cite{schapire2002incorporating,lefort2010weakly}, but its systematical integration into UDA solutions remains under-explored. 

\begin{figure}[!t]
	\begin{center}
		\centering
		\includegraphics[width=1.0\linewidth]{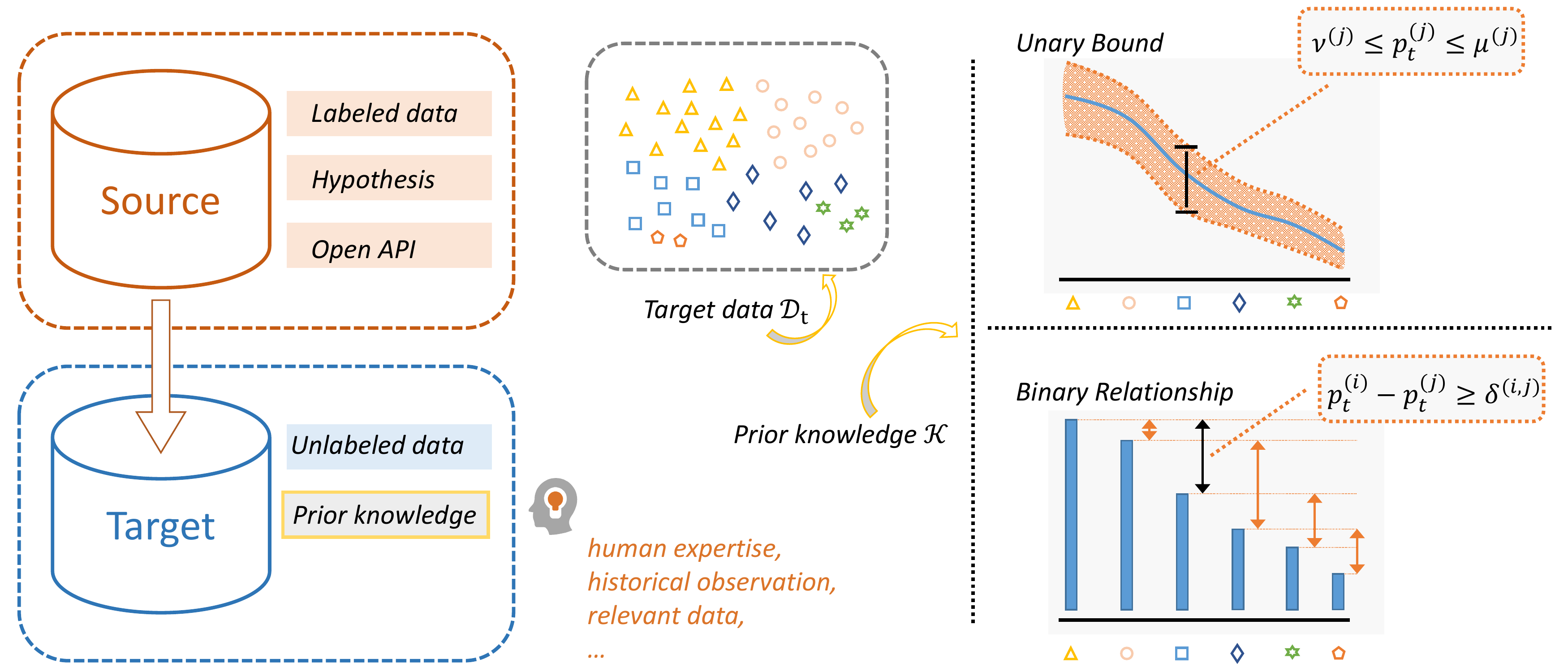}
	\end{center}
	\caption{(Left) Knowledge-guided Unsupervised Domain Adaptation (KUDA). In addition to target data, some prior knowledge about target class distribution is available. (Right) two types of prior knowledge considered in the paper. }		
	\label{fig:pk}
\end{figure}

Inspired by the above observation, in this paper we study a novel setting of UDA, named \emph{Knowledge-guided Unsupervised Domain Adaptation} (KUDA), as illustrated in Fig.~\ref{fig:pk}. Specifically, in addition to target training samples $\mathcal{D}_t$, a collection of prior knowledge $\mathcal{K}$ on target class distribution $p_t(y)$ is accessible. In particular, we consider two types of prior knowledge: \emph{Unary Bound} that describes the lower and upper bounds of individual class probability $p_t^{(c)}$ (\eg, the probability of ``square" is between 0.1 and 0.3), and \emph{Binary Relationship} that describes the relations between probabilities of two classes, $p_t^{(c_1)}$ and $p_t^{(c_2)}$ (\eg, there are more ``triangles" than ``squares"). The task of KUDA is to adapt a predictive model learned from a source domain to a target domain under the guidance from such prior knowledge.  It is worth mentioning that there can be many other types of prior knowledge which may help to improve UDA performance, and we choose unary and binary statistics over the class distribution for their generality and accessibility in practice.

To incorporate the prior knowledge into domain adaptation, we propose a novel \textit{rectification module} to refine model generated pseudo labels. We formulate the rectification procedure using prior knowledge as a \textit{Zero-One Programming} (ZOP)~\cite{wolsey2020integer} problem, where its optimal solution returns the updated pseudo labels. Moreover, smooth regularization is applied to maintain consistency of pseudo labels in neighboring samples. This module can be easily integrated into self-training-based UDA methods. To validate its effectiveness, we choose two recent state-of-the-art UDA methods, SHOT~\cite{liang2020we} and DINE~\cite{liang2021dine}, and improve them with the rectification module. 

The experimental validation is conducted on four commonly used UDA benchmarks, two of which have a large label distribution shift by design. The results confirm that the rectification module improves the quality of pseudo labels and hence benefits the self-training stage. Consequently, the performances of two methods under the guidance of prior knowledge, named respectively kSHOT and kDINE, are substantially boosted compared with the vanilla versions. Our work demonstrates that it is important to consider target class prior knowledge, especially when the domain gap is large. 

In summary, we make the following contributions:
\begin{itemize}
	\item We study a novel and practical setting of Knowledge-guided Unsupervised Domain Adaptation (KUDA), where prior knowledge about target class distribution available in addition to unlabeled training samples.
	\item We introduce a general rectification module that refines pseudo labels with the guidance from prior knowledge. It can be easily plugged into self-training based UDA methods.
	\item Extensive experiments on both standard and label-shifted benchmarks validate that incorporating prior knowledge can significantly boost the performance of adapted models, reducing the reliance on target training data.  
\end{itemize}

\section{Related Work}
\textbf{Incorporating Prior Knowledge.} There has been a long history of incorporating prior knowledge into machine learning tasks. Using prior knowledge removes or reduces the reliance on training data. The knowledge can be expressed in various forms, such as statistical descriptions from other data or human expertise, inductive biases, physical models, \etc. The most related one to our work is \emph{target prior}, where the distribution of target variable $p(y)$ is known~\cite{schapire2002incorporating,lefort2010weakly}. In~\cite{mann2007simple}, the class distribution prior conditioned on certain inputs is captured by generalized expectation. Zhu \etal~\cite{zhu2003semi} employ class priors to set thresholds on the propagation of labels. Wang \etal~\cite{wang2012learning} assume that a parametric target prior model $p_{\eta}(y)$ can be obtained from relevant subjects yet having no correspondence with training data. Inductive biases have been widely used in deep neural networks. A canonical one is translation equivariance through convolutions~\cite{kayhan2020translation,urban2017deep,lecun1989backpropagation}. Lin \etal~\cite{lin2020deep} add geometric priors based on Hough transform in line detection. Physical models of image formation have been integrated into the tasks of image decomposition~\cite{baslamisli2018cnn}, rain image restoration~\cite{li2019heavy}, day-night adaptation~\cite{lengyel2021zero}, \etc.

\noindent\textbf{Domain Adaptation Settings.} Domain Adaptation (DA) presents under many different settings. In the vanilla Unsupervised Domain Adaptation (UDA)~\cite{zhang2019bridging,prabhu2021sentry,chen2022reusing,sun2022safe}, only labeled data from a source domain and unlabeled data from a target domain are available. Since no labeled target data is available, UDA can be a challenging task when the domain gap is large. Semi-supervised DA (SSDA)~\cite{saito2019semi,kim2020attract,li2021learning} assumes a few labeled target data is available, which often greatly boosts performance compared with UDA. Active DA~\cite{fu2021transferable,prabhu2021active,xie2022active} further selects the most informative samples to query their labels from the oracle. Then human-defined criteria like uncertainty and diversity can be injected to measure the informativeness of samples. SSDA and Active DA can be viewed as incorporating additional instance-level label information compared with UDA. Another line is to reduce the information released by source domain, usually due to some privacy issues. Source-data free UDA~\cite{liang2020we,huang2021model,ding2022source,chen2022contrastive} assumes only a trained model is offered by the source domain while source data are inaccessible. To conceal model details, the black-box source model~\cite{lipton2018detecting,zhang2021unsupervised,liang2021dine} is further studied. 

\noindent\textbf{Our study.} Our proposed KUDA methods incorporate class distribution-level information and is complementary to all of the above-mentioned settings. It is the first work along this direction, and we expect to see further studies to explore richer prior knowledge for UDA or to extend the idea to general DA scenarios.

\section{Knowledge-guided UDA}

\noindent\textbf{Preliminaries.} In this paper, we focus on $C$-way classification problem for UDA tasks. We use $\mathcal{X}$ and $\mathcal{C}=\{0,1,\ldots,C-1\}$ to denote the input space and the label space respectively. In a vanilla UDA task, we are given labeled samples $\mathcal{D}_s=\{(\bm{x}_i^s, y_i^s)\}_{i=0}^{n_s-1}$ from a source domain $\mathcal{P}_{S}(\mathcal{X}, \mathcal{C})$, and unlabeled samples $\mathcal{D}_t=\{(\bm{x}_i^t)\}_{i=0}^{n_t-1}$ from a target domain $\mathcal{P}_{T}(\mathcal{X}, \mathcal{C})$. The goal of UDA is to learn a labeling function $f_t=h_t \circ g_t :\mathcal{X}\rightarrow \mathcal{C}$ for target domain, where $g_t$ is the feature extractor and $h_t$ is the label predictor.

\noindent\textbf{Prior Knowledge of Target Class Distribution.}
The class distribution of target domain, $p_t(y)$, is an important quantity while inaccessible in UDA. One way is to estimate it from model predictions on unlabeled target data~\cite{liu2021adversarial}. However, this can often be unreliable when the domain gap is large. The deficiency of labeled target samples can be compensated with prior knowledge, \eg, from human expertise. In fact, it is often possible to obtain some information about class distribution without bothering labeling  specific target samples in real-world applications. Table~\ref{tab:pk} lists two types of prior knowledge considered in this paper. \emph{Unary Bound} describes the lower and upper bounds of individual class probability $p_t^{(c)}$, and \emph{Binary Relationship} describes the relations between probabilities of two classes, $p_t^{(c_1)}$ and $p_t^{(c_2)}$. Both statistics over the class distribution are general and easy to obtain in practice. Other types of prior knowledge beyond these can be similarly defined in terms of three or more probabilities.

\begin{table}[!t]
	\begin{center}
		\caption{Two types of prior knowledge considered in the paper.}
		\label{tab:pk}
		\renewcommand\arraystretch{1.5}
		\begin{tabular}{cp{0.5cm}l}
			\toprule
			\textbf{Knowledge Type} && \textbf{Formulation}\\
			\midrule
			\textbf{Unary  Bound (UB)} &&  $\mathcal{K}=\left\{\left(\nu^{(c)} \leq p_t^{(c)} \leq \mu^{(c)}\right) \middle| c\in \mathcal{C}\right\} $ \\
			\textbf{Binary Relationship (BR)} &&  $\mathcal{K}=\left\{\left(p_t^{(c_1)} -p_t^{(c_2)} \geq \delta^{(c_1,c_2)} \right) \middle| c_1,c_2\in \mathcal{C}\right\} $\\
			\bottomrule
		\end{tabular}
	\end{center}
	
\end{table}

\noindent\textbf{Setting of KUDA.}
We study a novel and realistic setting termed \textbf{Knowledge-guided Unsupervised Domain Adaptation (KUDA)}. In KUDA, in addition to $\mathcal{D}_s$ and $\mathcal{D}_t$, we have access to some prior knowledge $\mathcal{K}$ about the target class distribution $p_t^{c}$. Such prior knowledge may provide valuable clues that are complementary to the unlabeled  training data, and can be especially beneficial when there exists a large distribution shift between source and target domains. In particular, we assume that $\mathcal{K}$ can be expressed in a collection of inequality constraints as listed in Tab.~\ref{tab:pk}. The goal is to learn an optimal target labeling function $f_t$ under the guidance from the prior knowledge $\mathcal{K}$.

\section{Method}
\subsection{Rectify Pseudo Labels with Prior Knowledge}\label{sec:pk}

Let us consider a general situation. Suppose we have a model predicted class probability matrix $P\in \mathbb{R}^{{n_t}\times C}$ of target data. The pseudo label of the $i$-th sample can be obtained via $\hat{y}_i^t=\arg\max \bm{p}_i$, where $\bm{p}_i$ is the $i$-th row of $P$. This procedure can be equivalently expressed in a more compact form using one-hot label representation $\bm{l}_i$ and its matrix form $L$ (\ie, $\bm{l}_i$ is the $i$-th row of $L$)

\begin{equation}\label{eq:pk_obj}
	\hat{L}=\mathop{\arg\max}_{L} \langle L,P \rangle, \quad
		{\rm s.t.}~ 
		\left\{ \begin{array}{rl}
        		\textstyle\sum_c L_{i,c}=1, \quad & \forall i\in [n_t]\\[10pt]
        		L_{i,c}\in\{0,1\}, & \forall c\in \mathcal{C}, i\in [n_t]
		\end{array} \right.
\end{equation}
where $\langle\cdot,\cdot\rangle$ is the inner product of two matrices, and $[n_t]\triangleq\{0,1,\ldots,n_t-1\}$. For the optimal solution $\hat{L}$ of Eq.~\ref{eq:pk_obj}, $\hat{L}_{i,\hat{y}_i^t}=1$ and $\hat{L}_{i,c}=0\ \forall c\neq \hat{y}_i^t$.

Without any prior knowledge, the optimal $\bm{l}_i$ is assigned independently for each target sample. The empirical class probability $\hat{p}_{t}^{(c)}=\sum_{i}L_{i,c}/{n_t}$ is expected to be close to $p_{t}^{(c)}$. However, this is often violated when the model predictions are noisy. When the prior knowledge $\mathcal{K}$ about $p_t$ is available, we can use it to rectify pseudo labels so that $\hat{p}_{t}$ is more compliant with $p_t$.

\noindent\textbf{Hard constraint form.} Given the inequalities listed in Tab.~\ref{tab:pk}, we plug in $\hat{p}_{t}$ and add the constraints to the optimization problem in Eq.~\ref{eq:pk_obj}. Then the optimization problem in hard constraint form can be formulated as:
\begin{itemize}
	\item \textbf{Unary Bound}. \begin{equation}\label{eq:pk_obj_hcon_ub}
	\hat{L}=\mathop{\arg\max}_{L} \langle L,P \rangle,
    \quad\text{s.t.}~ \left\{ 
		\begin{aligned}
		\ & \textstyle\sum_c L_{i,c}=1\quad \forall i\in [n_t]\\
		\ & L_{i,c}\in\{0,1\},\quad \forall c\in \mathcal{C}, i\in [n_t]\\
		\ & \textstyle\sum_{i}{L_{i,c}} \geq {n_t}\nu^{(c)}  ,\quad \forall c\in \mathcal{C}\\
		\ & -\textstyle\sum_{i}{L_{i,c}} \geq -{n_t}\mu^{(c)},\quad \forall c\in \mathcal{C}
		\end{aligned} \right.
\end{equation}
\item \textbf{Binary Relationship}.	
\begin{equation}\label{eq:pk_obj_hcon_br}
   \! \hat{L}=\mathop{\arg\max}_{L} \langle L,P \rangle, 
    ~\text{s.t.}~ \left\{ 
		\begin{aligned}
		\ & \textstyle\sum_c L_{i,c}=1\quad \forall i\in [n_t]\\
		\ & L_{i,c}\in\{0,1\},\quad \forall c\in \mathcal{C}, i\in [n_t]\\
		\ & \textstyle\sum_{i}{(L_{i,c_1} - L_{i,c_2})} \geq {n_t}\delta^{(c_1, c_2)},\quad \forall c_1,c_2 \in \mathcal{C}
		\end{aligned} \right.
\end{equation}
\end{itemize}

\noindent Eq.~\ref{eq:pk_obj_hcon_ub} and Eq.~\ref{eq:pk_obj_hcon_br} are \emph{Zero-One Programming} problems~\cite{wolsey2020integer}, and can be solved with standard solvers~\cite{gurobi}. However, using hard constraint form is not favored. When these constraints are inconsistent, the optimization problem becomes infeasible.

\noindent\textbf{Soft constraint form.} To overcome the drawbacks of hard constraint form, we convert prior knowledge into soft constraints by introducing slack variables:
\begin{itemize}
	\item \textbf{Unary Bound}. \begin{equation}\label{eq:pk_obj_scon_ub}
	\begin{aligned}
		&\quad\quad\hat{L}=\mathop{\arg\max}_{L} \langle L,P \rangle - M \sum_c (\xi_c^{(\nu)}+\xi_c^{(\mu)})\\
     	&	\text{s.t.}\quad \left\{ 
		\begin{aligned}
		\ & \textstyle\sum_c L_{i,c}=1\quad \forall i\in [n_t]\\
		\ & L_{i,c}\in\{0,1\},\quad \forall c\in \mathcal{C}, i\in [n_t]\\
		\ & \xi_c^{(\nu)}=\max\left(0,-\textstyle\sum_{i}{L_{i,c}} + {n_t}\nu^{(c)}\right),\quad \forall c\in \mathcal{C}\\
		\ & \xi_c^{(\mu)}=\max\left(0,\textstyle\sum_{i}{L_{i,c}} - {n_t}\mu^{(c)}\right),\quad \forall c\in \mathcal{C}
		\end{aligned} \right.
		\\
		\end{aligned} 
\end{equation}
\item \textbf{Binary Relationship}.	
\begin{equation}\label{eq:pk_obj_scon_br}
	\begin{aligned}
		&\quad\quad\hat{L}=\mathop{\arg\max}_{L} \langle L,P \rangle - M \sum_{c_1,c_2} \xi_{c_1,c_2}\\
     	&	\text{s.t.}\quad \left\{ 
		\begin{aligned}
		\ & \textstyle\sum_c L_{i,c}=1\quad \forall i\in [n_t]\\
		\ & L_{i,c}\in\{0,1\},\quad \forall c\in \mathcal{C}, i\in [n_t]\\
		\ & \xi_{c_1,c_2}=\max\left(0,-\textstyle\sum_{i}({L_{i,c_1} - L_{i,c_2}}\right) + {n_t}\delta^{(c_1, c_2)}),\quad \forall c_1,c_2 \in \mathcal{C}
		\end{aligned} \right.
		\\
		\end{aligned}
\end{equation}
\end{itemize}
 
In both Eq.~\ref{eq:pk_obj_scon_ub} and Eq.~\ref{eq:pk_obj_scon_br}, $M$ is a pre-defined non-negative constant. When $M$ is large enough, their solutions will be the same as those of Eq.~\ref{eq:pk_obj_hcon_ub} and Eq.~\ref{eq:pk_obj_hcon_br} respectively, providing the hard constraints from prior knowledge are satisfiable. When $M=0$, Eq.~\ref{eq:pk_obj_scon_ub} and Eq.~\ref{eq:pk_obj_scon_br} will degenerate to the vanilla problem in Eq.~\ref{eq:pk_obj}.

\noindent\textbf{Smooth regularization.} Previous optimization problems utilize prior knowledge about class distribution to refine pseudo labels. However, this solely relies on the model predicted probability matrix $P$ without considering the data distribution in the feature space. In classification tasks, it is expected that the label prediction is locally smoothed. Hence, we add a smooth regularization that enforces the pseudo label of neighboring samples to be consistent.

We select a subset of target samples $ \mathcal{S}_t \subseteq \mathcal{D}_t$ whose model predictions are uncertain. For each $\bm{x}^t_i\in \mathcal{S}_t$, let its nearest neighbor in $ \mathcal{D}_t \setminus \mathcal{S}_t$ be $\bm{x}^t_{ki}$. The smooth regularization is a collection of equality constraints, $\mathcal{R}=\{(\bm{l}_i=\bm{l}_{ki}) | \bm{x}^t_i\in \mathcal{S}_t\}$. Converting these equalities into soft constraints is non-trivial as it will bring second-order terms in the objective. Instead, we directly add them as hard constraints to the optimization problem in Eq.~\ref{eq:pk_obj_scon_ub} and Eq.~\ref{eq:pk_obj_scon_br}.  

\subsection{Knowledge-guided UDA Methods}
\begin{figure}[!t]
	\begin{center}
		\centering
		\includegraphics[width=0.97\linewidth]{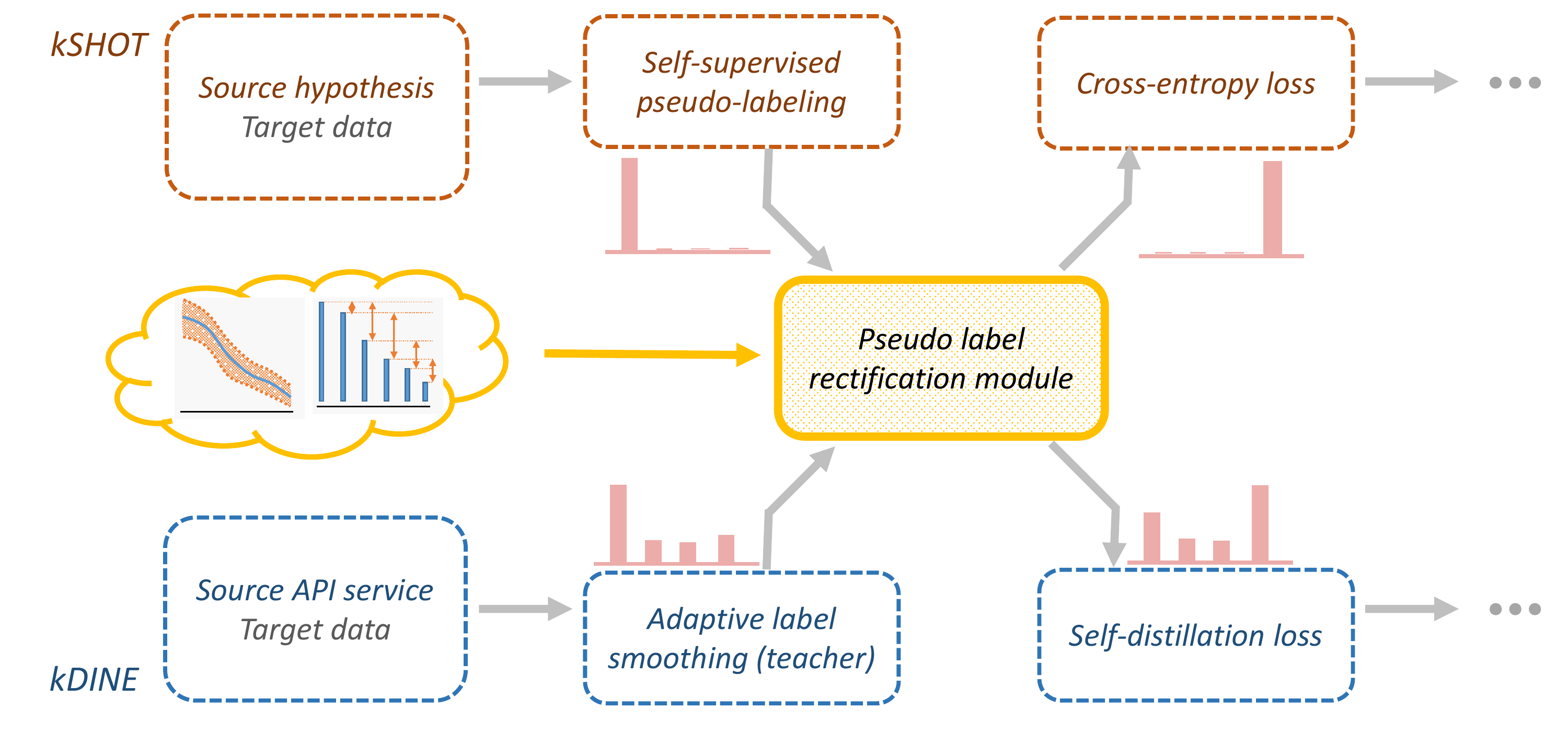}
	\end{center}
	\caption{Illustration of how our proposed rectification module is integrated into SHOT and DINE to get knowledge-guided SHOT and DINE. This can be easily extended to other self-training based UDA methods in a similar manner.}  
	\label{fig:framework}
\end{figure}

Our proposed rectification module is general, and can be easily plugged into self-training based UDA methods. To validate its effectiveness, we choose two recent UDA algorithms, SHOT~\cite{liang2020we} and DINE~\cite{liang2021dine}. This leads to knowledge-guided SHOT and DINE, dubbed as kSHOT and kDINE, respectively. The frameworks are illustrated in Fig.~\ref{fig:framework}. It should be noted that the main purpose of this part is to demonstrate the benefits of considering class prior knowledge through our rectification module, rather than simply extending the two algorithms.  

\noindent\textbf{Knowledge-guide SHOT.} SHOT~\cite{liang2020we} is a state-of-the-art self-training based UDA method. It assumes a source-data free setting, \ie, only the source hypothesis is available during adaptation. Then  both self-supervised pseudo-labeling and mutual information maximization are exploited to fine-tune the feature extractor module of the source hypothesis. Since only target samples are involved, it provides us a convenient platform to observe how prior knowledge affects the performance of adapted models. The full objective is
\begin{equation}\label{eq:shot_obj}
	\mathcal{L}_{\rm shot}=\mathbb{E}_{(\bm{x}^t_i, \hat{y}^t_i)} \ell_{\rm ce}(h_t\circ g_t (\bm{x}^t_i), \hat{y}^t_i)- \alpha \mathcal{L}_{\rm im} 
\end{equation} 
where $\ell_{\rm ce}$ is the cross entropy loss, $\mathcal{L}_{\rm im}$ is the Information Maximization loss~\cite{gomes2010discriminative,shi2012information}, and $\alpha$ is a hyper-parameter. A critical step is to obtain the pseudo label $\hat{y}^t_i$. SHOT uses the distances between samples and class centroids in the feature space to refine model predictions. While this improves the quality of pseudo labels to some extent, their empirical distribution could still be very different from the ground-truth, as shown in Fig.~\ref{fig:hist}. We show how prior knowledge can be incorporated to alleviate this issue. 

We plug the rectification module into SHOT. After obtaining pseudo labels $L^{({\rm shot})}$ (\ie, the one-hot representation of $\hat{y}^t$) and feature-to-centroid distances from SHOT, we convert the distances into class probabilities using \texttt{softmax} as
\begin{equation}
	P = \texttt{softmax}(-D)
\end{equation}
where $D\in \mathbb{R}^{{n_t}\times C}$, $D_{i,k}=d_f( g_t(\bm{x}_i^t), c_{k})$, $d_f$ is some distance metric and $c_k$ is the $k$-th class centroid.  

Let the optimal solution of the Zero-One Programming in Eq.~\ref{eq:pk_obj_scon_ub} and Eq.~\ref{eq:pk_obj_scon_br} under prior knowledge $\mathcal{K}$ and smooth regularization $\mathcal{R}$ be
\begin{equation}
	L^{*} = \mathfrak{S}(P,\mathcal{K}, \mathcal{R})
\end{equation}
$\mathfrak{S}(P,\emptyset, \emptyset)$ returns exactly the same pseudo labels as $L^{({\rm shot})}$. Given prior knowledge $\mathcal{K}$, we first obtain $L^{(\rm pk_0)}=\mathfrak{S}(P,\mathcal{K}, \emptyset)$. Then we create a subset $\mathcal{S}_t=\{x^t_i|\bm{l}^{({\rm shot})}_i\neq\bm{l}^{(\rm pk_0)}_i\}$ that consists of all samples whose pseudo label changed. After that the smooth regularization $\mathcal{R}$ can be constructed using $\mathcal{S}_t$. Finally, we obtain $L^{(\rm pk_1)}=\mathfrak{S}(P, \mathcal{K}, \mathcal{R})$ and use $L^{(\rm pk_1)}$ as $\hat{y}^t$ to update model with Eq.~\ref{eq:shot_obj}.

\noindent\textbf{Knowledge-guide DINE.} DINE~\cite{liang2021dine} is a very recent algorithm that assumes only black-box source models (\eg, source API service) are available during adaptation. It first distills knowledge from the source predictor to a target model, and then fine-tunes the distilled model with target data. Two kinds of structural regularization, including interpolation consistency training~\cite{zhang2017mixup} and mutual information maximization~\cite{shi2012information}, are applied. The objective is 
\begin{equation}\label{eq:dine_obj}
	\mathcal{L}_{\rm dine}= \mathbb{E}_{\bm{x}^t_i}\mathcal{D}_{\rm kl}\left(P^{\rm tch}(\bm{x}^t_i)\|f_t(\bm{x}^t_i)\right) + \beta \mathcal{L}_{\rm mix} - \mathcal{L}_{\rm im}
\end{equation} 
where $\mathcal{D}_{\rm kl}$ denotes the Kullback-Leibler divergence, $\mathcal{L}_{\rm mix}$ and $\mathcal{L}_{\rm im}$ are two regularizers, and $\beta$ is a trade-off parameter. To obtain the teacher prediction $P^{\rm tch}(\bm{x}^t_i)$, the authors propose to revise the predictions of source model with adaptive label smoothing, and maintain an exponential moving average (EMA) prediction. 

Given prior knowledge $\mathcal{K}$, we aim to rectify the teacher prediction $P^{\rm tch}(\bm{x}^t_i)$ to be more compliant with the ground-truth. We adopt a similar strategy as in kSHOT. The pseudo labels of DINE are obtained by $\hat{y}^{({\rm dine})}_i=\arg\max P^{\rm tch}(\bm{x}^t_i)$. Let the corresponding one-hot representation be $\bm{l}^{({\rm dine})}_i$. We take $P^{\rm tch}(\bm{x}^t_i)$ as the $i$-th row of $P$, and obtain $L^{(\rm pk_0)}=\mathfrak{S}(P, \mathcal{K}, \emptyset)$. Then we create a subset $\mathcal{S}^t=\{\bm{x}^t_i|\bm{l}^{({\rm dine})}_i\neq\bm{l}^{(\rm pk_0)}_i\}$ to construct the smooth regularization $\mathcal{R}$. Finally, we obtain $L^{(\rm pk_1)}=\mathfrak{S}(P, \mathcal{K}, \mathcal{R})$. The new objective function is

\begin{equation}\label{eq:dine_obj_pk}
	\mathcal{L}_{\rm kdine}= \mathbb{E}_{\bm{x}^t_i}\mathcal{D}_{\rm kl}\left(\frac{P^{\rm tch}(\bm{x}^t_i)+ \tilde{\bm{l}}_i^{(\rm pk_1)}}{2}  \middle\| f_t(\bm{x}^t_i)\right) + \beta \mathcal{L}_{\rm mix} - \mathcal{L}_{\rm im}
\end{equation} 
where $\tilde{\bm{l}}_i^{(\rm pk_1)}=0.9\cdot\bm{l}_i^{(\rm pk_1)}+0.1/C$ is the smoothed label.

\section{Experiments}
\subsection{Experimental Setup}\label{sec:exp_setup}

\textbf{Datasets.} We report our results on both standard UDA benchmarks and benchmarks designed with label distribution shift. \textbf{Office-Home} is an image classification dataset with 65 classes from four environments: Artistic (A), Clip Art (C), Product (P), and Real-world (R). \textbf{Office}~\cite{saenko2010adapting} contains 31 classes of office objects from three domains: Amazon (A), DSLR (D) and Webcam (W).  \textbf{VisDA-2017}~\cite{peng2017visda} is a large-scale Synthetic-to-Real dataset with 12 categories of objects. \textbf{Office-Home RS-UT}~\cite{tan2019generalized} is a subset of Office-Home created with Reverse-unbalanced Source and Unbalanced Target manner. Both source and target label distributions are long-tailed. The majority classes in source domain are minority ones in target domain. Hence, it has a big label distribution shift.  \textbf{DomainNet}~\cite{peng2019moment} is a large UDA benchmark. We use the subset~\cite{tan2019generalized} of 40-commonly seen classes from four domains: Clipart (C), Painting (P), Real (R), Sketch (S). It has a natural label distribution shift.

\noindent\textbf{Creating prior knowledge.} We create prior knowledge from ground-truth labels of target training data,  $\{y_i^t\}_{i=0}^{n_t-1}$, for experimental purposes only. The noisiness and completeness of the prior knowledge are discussed in Sec.~\ref{sec:analysis}. Let $q_c=\sum_{i}\mathbb{I}[y^t_i=c]/n_t$ be the empirical probability of the $c$-th class. 
\begin{itemize}
	\item \textbf{Unary Bound}. We create UB as  $\big\{(q_c\cdot (1-\sigma) \leq p_t^{(c)} \leq q_c\cdot (1+\sigma)) | c\in \mathcal{C}\big\}$, where $\sigma$ is hyper-parameter controlling the tightness of the bounds. In the experiments, we choose $\sigma\in\{0.0, 0.1, 0.5, 1.0, 2.0\}$.
	\item \textbf{Binary Relationship}. We first sort all classes based on $q_c$ in descending order. Assuming the corresponding indexes are $[c_0^q, c_1^q,\cdots,c_{C-1}^q]$. Then we create BR as $\big\{(p_t^{(c_i^q)} -p_t^{(c_{i+1}^q)} \geq 0 ) | i\in\{0,1,\cdots,C-2\}  \big\}$. We simply take the right hand to be 0, which makes them relatively loose constraints and more easily available in practice.
\end{itemize}

\noindent\textbf{Implementation details.}
We solve the optimization problem of the rectification module with Gurobi Optimizer~\cite{gurobi}. Both kSHOT and kDINE are based on the official Pytorch implementations by the authors. We use a pretrained ResNet-101~\cite{he2016deep} backbone for VisDA-2017, and ResNet-50~\cite{he2016deep} for others. To fairly compare with SHOT and DINE, we adopt the same hyper-parameters as used in the original papers. We run every task for 3 times and report the mean evaluation values. For standard UDA benchmarks, we report the accuracy. For benchmarks with label distribution shift (\ie, Office-Home RS-UT and DomainNet), we report per-class average accuracy, in consistent with previous works~\cite{tan2019generalized,prabhu2021sentry}. 

\begin{table}[!t]
	\centering
	\caption{Classification accuracies (\%) on \textbf{Office-Home RS-UT} and \textbf{Office}.}
	\scalebox{0.88}{
		\begin{tabular}{p{1.0cm}@{}p{0.6cm}<{\centering}@{}p{0.6cm}<{\centering}|@{}p{0.8cm}<{\centering}@{}p{0.8cm}<{\centering}@{}p{0.8cm}<{\centering}@{}p{0.8cm}<{\centering}@{}p{0.8cm}<{\centering}@{}p{0.8cm}<{\centering}@{}>{\columncolor{tbgray}[-0.2ex][0.3ex]}p{0.8cm}<{\centering}|@{}p{0.8cm}<{\centering}@{}p{0.8cm}<{\centering}@{}p{0.8cm}<{\centering}@{}p{0.8cm}<{\centering}@{}p{0.8cm}<{\centering}@{}p{0.8cm}<{\centering}@{}>{\columncolor{tbgray}[-0.2ex][0.3ex]}p{0.8cm}<{\centering}}
			\toprule
			\multirow{2}{*}{Method} &  \multirow{2}{*}{$\mathcal{K}$} & \multirow{2}{*}{$\sigma$}& \multicolumn{7}{c|}{Office-Home RS-UT} & \multicolumn{7}{c}{Office}\\
			 &  &  & R$\shortrightarrow$P &	R$\shortrightarrow$C &	P$\shortrightarrow$R &	P$\shortrightarrow$C &	C$\shortrightarrow$R & C$\shortrightarrow$P & Avg. &  A$\shortrightarrow$D & A$\shortrightarrow$W & D$\shortrightarrow$A & D$\shortrightarrow$W & W$\shortrightarrow$A & W$\shortrightarrow$D & Avg.    \\ 	
			\midrule
			SHOT & -- & -- & 77.0 &	50.3 &	75.9 &	47.0 &	64.3 &	64.6 &	63.2 & 94.0 &	90.1 &	74.7 &	98.4 &	74.3 &	\HL{99.9} &	88.6 		
			\\
			\midrule
			\multirow{6}{*}{\rotatebox[origin=c]{90}{kSHOT}} & UB & 0.0 &\HL{78.8} &	51.3 &	\HL{79.1} &	\HL{49.8} &	71.3 &	\HL{69.7} &	\HL{66.6} &  \HL{97.6} &	\HL{98.5} &	75.0 &	\HL{99.0} &	76.2 &	99.8 &	\HL{91.0} \\
			& UB & 0.1 & 78.3 &	\HL{51.7} &	79.0 &	48.6 &	\HL{71.6} &	69.4 &	66.4 & 96.7 &	97.2 &	75.5 &	98.7 &	\HL{76.5} &	99.8 &	90.7 \\
			& UB & 0.5 & 76.7 &	50.6 &	77.3 &	48.4 &	68.4 &	67.3 &	64.8  & 93.9 &	92.8 &	\HL{75.7} &	97.7 &	75.2 &	99.7 &	89.2 	\\
			& UB & 1.0 & 76.3 &	50.0 &	76.9 &	48.4 &	66.7 &	65.8 &	64.0 & 93.7 &	92.4 &	75.5 &	97.7 &	75.5 &	99.7 &	89.1 	\\
			& UB & 2.0 & 76.4 &	50.0 &	76.0 &	47.9 &	65.3 &	64.1 &	63.3 & 93.7 &	92.4 &	75.0 &	97.7 &	75.2 &	99.7 &	89.0 	 \\
			& BR & -- & 78.6 &	51.6 &	78.7 &	49.3 &	70.1 &	68.8 &	66.2 & 96.9 &	97.1 &	74.0 &	98.8 &	76.1 &	99.8 &	90.5  \\
			\bottomrule
		\end{tabular} 
	}
	\label{tab:officehome-rsut_office}
\end{table}

\begin{table}[!t]
	\caption{Classification accuracies (\%) on \textbf{Office-Home} and \textbf{VisDA-2017}.}
	\scalebox{0.88}{
	\centering
	\begin{tabular}{p{1.0cm}@{}p{0.6cm}<{\centering}@{}p{0.6cm}<{\centering}|@{}p{0.8cm}<{\centering}@{}p{0.8cm}<{\centering}@{}p{0.8cm}<{\centering}@{}p{0.8cm}<{\centering}@{}p{0.8cm}<{\centering}@{}p{0.8cm}<{\centering}@{}p{0.8cm}<{\centering}@{}p{0.8cm}<{\centering}@{}p{0.8cm}<{\centering}@{}p{0.8cm}<{\centering}@{}p{0.8cm}<{\centering}@{}p{0.8cm}<{\centering}@{}>{\columncolor{tbgray}}p{0.8cm}<{\centering}|p{0.8cm}<{\centering}}
		\toprule
		\multirow{2}{*}{Method} &  \multirow{2}{*}{$\mathcal{K}$} & \multirow{2}{*}{$\sigma$}& \multicolumn{13}{c|}{Office-Home} & \multirow{2}{*}{VisDA}\\
	  & & & A$\shortrightarrow$C & A$\shortrightarrow$P & A$\shortrightarrow$R & C$\shortrightarrow$A & C$\shortrightarrow$P &  C$\shortrightarrow$R &  P$\shortrightarrow$A & P$\shortrightarrow$C & P$\shortrightarrow$R & R$\shortrightarrow$A & R$\shortrightarrow$C & R$\shortrightarrow$P & Avg. &   \\ 	
		\midrule
	    SHOT &-- & -- & 57.1 &	78.1 &	81.5 &	68.0 &	78.2 &	78.1 &	67.4 &	54.9 &	82.2 &	73.3 &	58.8 &	84.3 &	71.8 & 82.9
		\\
		\midrule
        \multirow{6}{*}{\rotatebox[origin=c]{90}{kSHOT}} & UB & 0.0 & \HL{58.2} &	\HL{80.0} &	82.9 &	\HL{71.1} &	\HL{80.3} &	\HL{80.7} &	71.3 &	\HL{56.8} &	\HL{83.2} &	75.5 &	60.3 &	\HL{86.6} &	\HL{73.9} & \HL{86.1}\\	
         & UB & 0.1 & 58.1 &	79.2 &	\HL{83.2} &	70.4 &	80.0 &	\HL{80.7} &	\HL{71.4} &	56.5 &	83.0 &	\HL{75.6} &	\HL{60.8} &	86.0 &	73.7 & 85.8 \\
         & UB & 0.5 & 57.4 &	79.1 &	82.1 &	69.4 &	78.1 &	79.5 &	69.3 &	55.2 &	81.8 &	74.0 &	60.2 &	85.1 &	72.6 & 83.9\\
         & UB & 1.0 & 57.0 &	79.0 &	82.1 &	68.6 &	77.8 &	79.3 &	68.4 &	55.1 &	81.7 &	73.5 &	59.3 &	84.8 &	72.2 & 83.0\\
         & UB & 2.0 & 56.4	 & 78.7 &	82.1 &	68.3 &	77.8 &	79.3 &	67.9 &	54.2 &	81.7 &	73.3 &	58.7 &	84.8 &	71.9 & 82.6\\	
         & BR & -- & 57.4 &	78.8 &	82.9 &	70.7 &	80.0 &	80.5 &	70.8 &	55.0 &	82.8 &	74.6 &	59.9 &	86.0 &	73.3 & 83.6\\	
		\bottomrule
	\end{tabular} 
}
	\label{tab:officehome_visda}
\end{table}

\begin{table}[!t]
	\caption{Classification accuracies (\%) on \textbf{DomainNet}.}	
	\centering
	\scalebox{0.88}{
		\begin{tabular}{p{1.2cm}@{}p{0.6cm}<{\centering}@{}p{0.6cm}<{\centering}|@{}p{0.8cm}<{\centering}@{}p{0.8cm}<{\centering}@{}p{0.8cm}<{\centering}@{}p{0.8cm}<{\centering}@{}p{0.8cm}<{\centering}@{}p{0.8cm}<{\centering}@{}p{0.8cm}<{\centering}@{}p{0.8cm}<{\centering}@{}p{0.8cm}<{\centering}@{}p{0.8cm}<{\centering}@{}p{0.8cm}<{\centering}@{}p{0.8cm}<{\centering}@{}>{\columncolor{tbgray}}p{0.8cm}<{\centering}}
			\toprule
			Method & $\mathcal{K}$ & $\sigma$  & R$\shortrightarrow$C &	R$\shortrightarrow$P &	R$\shortrightarrow$S &	C$\shortrightarrow$R &	C$\shortrightarrow$P &	C$\shortrightarrow$S &	P$\shortrightarrow$R &	P$\shortrightarrow$C &	P$\shortrightarrow$S &	S$\shortrightarrow$R &	S$\shortrightarrow$C &	S$\shortrightarrow$P	 & Avg.   \\ 	
			\midrule
			SHOT & -- & -- & 79.4	 & 75.4 &	72.8 &	88.4 &	74.0 &	75.5 &	89.8 &	77.7 &	76.2 &	88.3 &	\HL{80.5} &	70.8 &	79.1		
			\\
			\midrule
			\multirow{6}{*}{\rotatebox[origin=c]{90}{kSHOT}} & UB & 0 & \HL{83.6} &	77.5 &	\HL{75.3} &	\HL{91.5} &	76.4 &	\HL{77.0} &	\HL{91.7} &	\HL{82.3} &	76.3 &	\HL{89.7} &	80.2 &	70.3 &	\HL{81.0}
			\\
			& UB & 0.1 & 82.2 &	\HL{77.6} &	75.2 &	89.5 &	\HL{76.8} &	76.9 &	91.2 &	81.7 &	\HL{76.9} &	88.5 &	79.4 &	70.1 &	80.5
			\\
			& UB & 0.5 & 80.3 &	77.2 &	73.5 &	88.8 &	75.4 &	75.6 &	89.0 &	78.4 &	76.6 &	88.3 &	78.9 &	70.4 &	79.4
			\\
			& UB & 1.0 & 79.9 &	76.8 &	72.9 &	88.8 &	73.7 &	75.3 &	88.6 &	77.6 &	76.4 &	88.0 &	80.0 &	69.9 &	79.0 \\
			& UB & 2.0 & 79.2 &	76.3 &	73.1 &	88.8 &	75.4 &	75.5 &	88.6 &	77.8 &	76.2 &	87.9 &	80.1 &	\HL{71.1} &	79.2 \\
			& BR & -- & 82.1 &	76.8 &	74.3 &	89.1 &	73.7 &	76.4 &	\HL{91.7} &	80.6 &	75.9 &	88.8 &	79.1 &	70.2 &	79.9 \\		
			\bottomrule
		\end{tabular} 
	}
	\label{tab:domainnet}
\end{table}

\subsection{Results}

\noindent\textbf{Results of kSHOT.}
Tables~\ref{tab:officehome-rsut_office},\ref{tab:domainnet} list results of kSHOT on two benchmarks with label distribution shift. In UB($\sigma=0$), it improves the accuracy by +3.4\% on Office-Home RS-UT and +1.9\% on DomainNet. As $\sigma$ grows, the prior knowledge becomes less informative, and consequently the improvements reduce. Interestingly in BR where only the relative order of class probabilities is known, it still improves +3.0\% on Office-Home RS-UT. Since this dataset is manually created to be long-tailed, and class distributions of two domains are reversed version of each other, having prior knowledge about target class distribution would be very helpful. This conforms to our experimental results. Results on three standard benchmarks are listed in Tables~\ref{tab:officehome-rsut_office},\ref{tab:officehome_visda}. Using prior knowledge consistently improves on them. Similar trends on $\sigma$ can be observed. Compared with previous benchmarks, the phenomenon of label distribution shift is less severe. Still prior knowledge can be helpful to correct mistaken pseudo labels during training.  

\begin{table}[!t]
\begin{minipage}{0.39\textwidth}
	\caption{Classification accuracies (\%) on \textbf{Office-Home} for PDA.}	
	\centering
	\scalebox{0.86}{
		\begin{tabular}{p{1.0cm}p{0.35cm}<{\centering}p{0.6cm}<{\centering}|p{0.6cm}<{\centering}p{0.6cm}<{\centering}p{0.6cm}<{\centering}p{0.6cm}<{\centering}>{\columncolor{tbgray}}p{0.6cm}<{\centering}}
		\toprule
		Method & $\mathcal{K}$ & $\sigma$  & :A & :C & :P & :R & Avg.   \\ 	
		\midrule
		SHOT & -- & -- & 78.9 &	65.2 &	82.9 &	90.3 &	79.3 \\
		\midrule
		\multirow{2}{*}{kSHOT} & UB & 0.0 & \HL{85.4} &	\HL{74.1} &	\HL{94.2} &	\HL{93.6} &	\HL{86.8}  \\
		& BR & -- & 84.9 &	72.3 &	90.2 &	92.2 &	84.9 \\	
		\midrule
		DINE & -- & -- & 77.6 &	59.2 &	82.7 &	85.2 &	76.2 \\
		DINE$^*$ & -- & -- &  73.1 &	54.8 &	80.0 &	83.9 &	73.0 \\
		\midrule
		\multirow{2}{*}{kDINE} & UB & 0.0 & \HL{82.1} &	\HL{66.4} &	\HL{91.3} &	\HL{91.7} &	\HL{82.9} \\
		& BR & -- & 79.7 &	63.3 &	88.2 &	89.5 &	80.2 \\
		\bottomrule
	\end{tabular} }
	\label{tab:officehome:pda}
\end{minipage}
\hfill
\begin{minipage}{0.58\textwidth}
	\caption{Classification accuracies (\%) on \textbf{Office}.} 
	\centering	
	\scalebox{0.86}{
	\renewcommand\arraystretch{1.1}
		\begin{tabular}{p{1.0cm}@{}p{0.6cm}<{\centering}@{}p{0.6cm}<{\centering}|@{}p{0.8cm}<{\centering}@{}p{0.8cm}<{\centering}@{}p{0.8cm}<{\centering}@{}p{0.8cm}<{\centering}@{}p{0.8cm}<{\centering}@{}p{0.8cm}<{\centering}@{}>{\columncolor{tbgray}[-0.2ex][0.35ex]}p{0.8cm}<{\centering}}
		\toprule
		Method & $\mathcal{K}$ & $\sigma$  & A$\shortrightarrow$D & A$\shortrightarrow$W & D$\shortrightarrow$A & D$\shortrightarrow$W & W$\shortrightarrow$A & W$\shortrightarrow$D & Avg. \\ 
		\midrule
		DINE & -- & -- &  91.6 & 86.8 & \HL{72.2} & 96.2 & \HL{73.3} & 98.6 & 86.4\\
		DINE$^*$ & -- & -- &  90.6 &	86.5 &	70.6 &	95.2 &	72.0 &	99.3 &	85.7 \\
		\midrule
		\multirow{6}{*}{\rotatebox[origin=c]{90}{kDINE}} & UB & 0.0  & \HL{94.7}  &	\HL{92.2}  &	71.0  &	\HL{96.8}  &	72.6  &	99.8  &	\HL{87.9}\\
		& UB & 0.1  & 93.6  &	91.2  &	71.0  &	96.5  &	72.1  &	\HL{99.5}  &	87.3   \\
		& UB & 0.5 & 91.7  &	88.3  &	70.4  &	95.2  &	71.6  &	99.3  &	86.1 \\
		& UB & 1.0 & 	90.6  &	86.4  &	70.6  &	95.2  &	72.3  &	99.3  &	85.8 \\	
     	& UB & 2.0 & 	90.6  &	86.5  &	70.7  &	95.2  &	72.0  &	99.3  &	85.7 \\
		& BR & -- &  93.4  &	91.1  &	70.5  &	96.4  &	72.1  &	\HL{99.5}  &	87.2 \\		
		\bottomrule
	\end{tabular}
}
	\label{tab:office_dine}
\end{minipage}

\end{table}

\begin{table}[!t]
	\caption{Classification accuracies (\%) on \textbf{Office-Home}.}	
	\centering
	\scalebox{0.88}{
	\begin{tabular}{p{1.4cm}@{}p{0.6cm}<{\centering}@{}p{0.6cm}<{\centering}|@{}p{0.8cm}<{\centering}@{}p{0.8cm}<{\centering}@{}p{0.8cm}<{\centering}@{}p{0.8cm}<{\centering}@{}p{0.8cm}<{\centering}@{}p{0.8cm}<{\centering}@{}p{0.8cm}<{\centering}@{}p{0.8cm}<{\centering}@{}p{0.8cm}<{\centering}@{}p{0.8cm}<{\centering}@{}p{0.8cm}<{\centering}@{}p{0.8cm}<{\centering}@{}>{\columncolor{tbgray}}p{0.8cm}<{\centering}}
			\toprule
			Method & $\mathcal{K}$ & $\sigma$ & A$\shortrightarrow$C & A$\shortrightarrow$P & A$\shortrightarrow$R & C$\shortrightarrow$A & C$\shortrightarrow$P & C$\shortrightarrow$R &  P$\shortrightarrow$A & P$\shortrightarrow$C & P$\shortrightarrow$R & R$\shortrightarrow$A & R$\shortrightarrow$C & R$\shortrightarrow$P & Avg.   \\ 	
			\midrule
			DINE & -- & -- & 52.2 & 78.4 & 81.3 & 65.3 & 76.6 & 78.7 & 62.7 & 49.6 & 82.2 & 69.8 & 55.8 & 84.2 & 69.7	\\
			DINE$^{*}$ & -- & --  & 51.8	 &76.0 &	79.6 &	63.1 &	75.1 &	76.5 &	60.4 &	48.5 &	80.7 &	69.4 &	55.9 &	83.5 &	68.4  \\
			\midrule
			\multirow{6}{*}{kDINE} & UB & 0.0 & 54.8 &	78.6 &	\HL{81.7} &	\HL{67.1} &	78.3 &	\HL{79.6} &	\HL{66.8} &	\HL{52.3} &	82.5 &	\HL{72.0} &	\HL{58.1} &	\HL{85.4} &	\HL{71.4} \\	
			& UB & 0.1 & \HL{55.0} &	78.8 &	81.1 &	66.4 &	77.7 &	79.2 &	66.4 &	51.8 &	82.3 &	71.5 &	58.0 &	84.9 &	71.1 \\
			& UB & 0.5 & 52.9 &	76.7 &	79.9 &	64.5 &	76.3 &	77.8 &	63.8 &	51.0 &	80.9 &	70.5 &	57.1 &	84.2 &	69.6 \\
			& UB & 1.0 & 52.3 &	76.0 &	79.6 &	63.5 &	75.2 &	76.5 &	62.1 &	49.0 &	80.7 &	69.9 &	56.4 &	83.4 &	68.7  \\
			& UB & 2.0 & 51.8 &	76.0 &	79.6 &	63.0 &	75.1 &	76.5 &	60.8 &	49.2 &	80.7 &	69.6 &	55.5 &	83.5 &	68.4 \\	
			& BR & -- &  54.2 &	\HL{79.4} &	81.5 &	66.8 &	\HL{78.6} &	79.2 &	65.6 &	50.9 &	\HL{82.6} &	71.4 &	\HL{58.1} &	85.3 &	71.1 \\	
			\bottomrule
		\end{tabular} 
	}
	\label{tab:officehome_dine}
\end{table}

\noindent\textbf{Results of kDINE.} Since using vanilla label smoothing is sub-optimal to the adaptive label smoothing used in DINE~\cite{liang2021dine}, the true performance gain from prior knowledge could be reduced. To make a fair comparison, we also provide results when replacing $\bm{l}_{i}^{(\rm pk_1)}$ with the $\bm{l}_{i}^{({\rm dine})}$ in Eq.~\ref{eq:dine_obj_pk}, and term it as DINE$^*$. As can be seen in Tables~\ref{tab:office_dine},\ref{tab:officehome_dine}, DINE$^*$ indeed performs worse than DINE. Nevertheless, incorporated with prior knowledge, kDINE achieves much higher accuracy, and even better performance than DINE. 

\noindent\textbf{Results for PDA.} We further evaluate on Office-Home for Partial-set DA (PDA), where there are totally 25 classes (the first 25 classes in alphabetical order) in the target domain and 65 classes in the source domain. This thus can be viewed as an extreme situation where class probabilities of the rest 40 classes are all zero. Table~\ref{tab:officehome:pda} lists the results averaged over tasks with the same target domain (\eg, the :A column averages over C$\rightarrow$A, P$\rightarrow$A and R$\rightarrow$A). As can be seen, using prior knowledge significantly improves in this situation.

\subsection{Analysis}\label{sec:analysis}

\begin{figure}[!t]
	\begin{center}
	\begin{minipage}{0.18\linewidth}
		\subfigure{
			\includegraphics[width=1.0\linewidth]{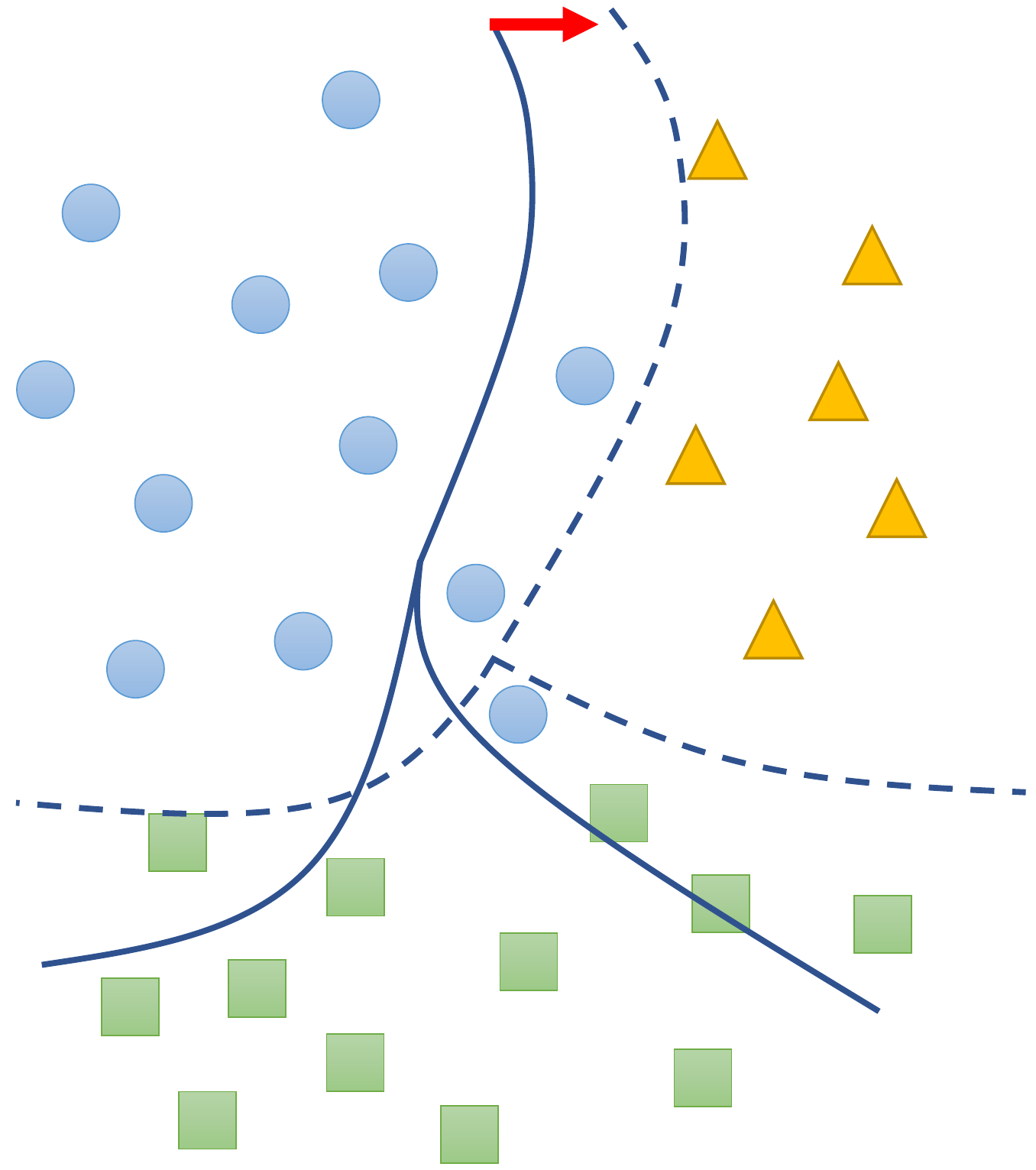} \label{fig:boundary} }
		\end{minipage}
	    \quad
		\begin{minipage}{0.35\linewidth}
		\subfigure{
			\includegraphics[width=1.0\linewidth]{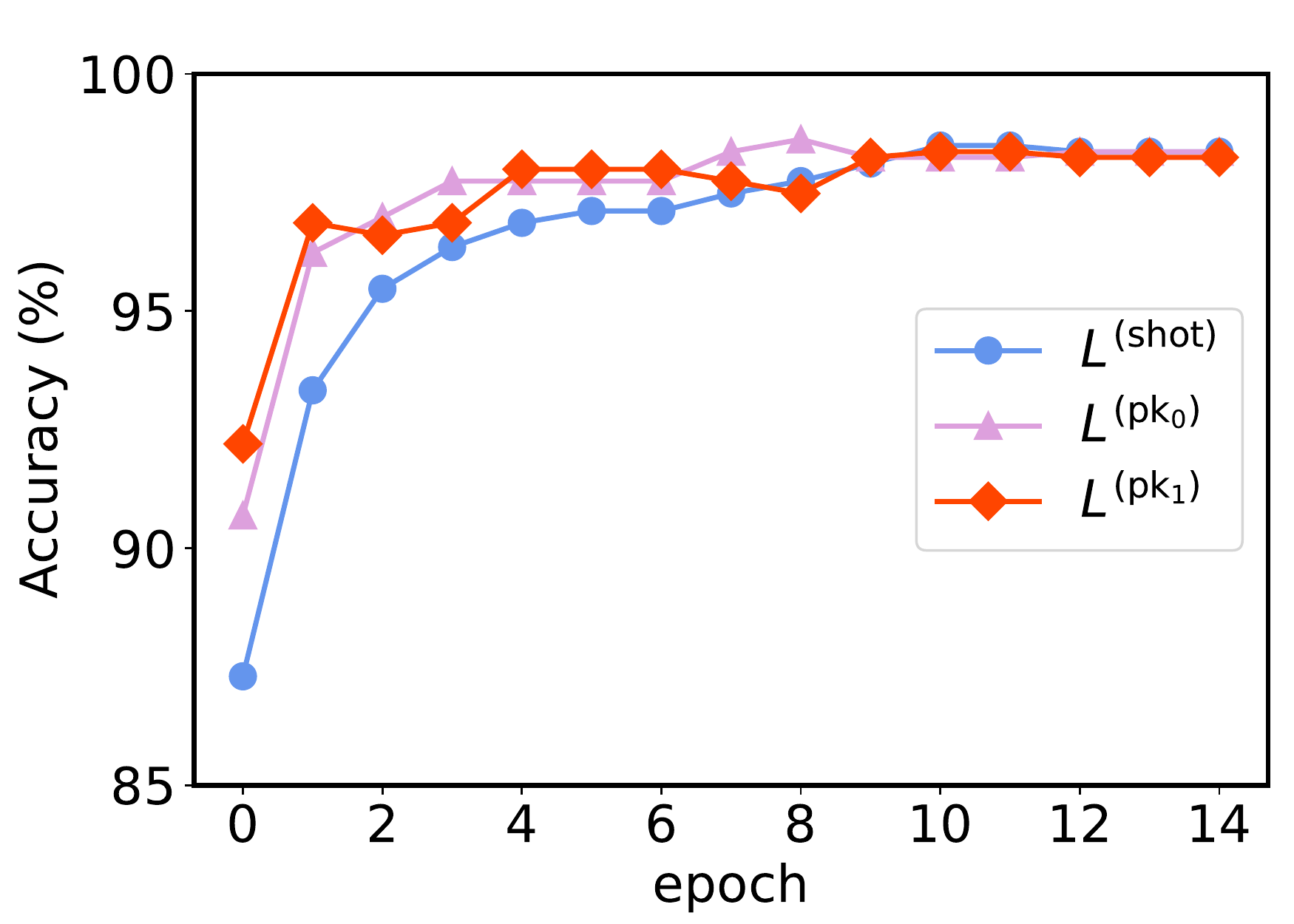} \label{fig:acc_curve}}
		\end{minipage}
		\quad
		\begin{minipage}{0.35\linewidth}
		\subfigure{
		\includegraphics[width=1.0\linewidth]{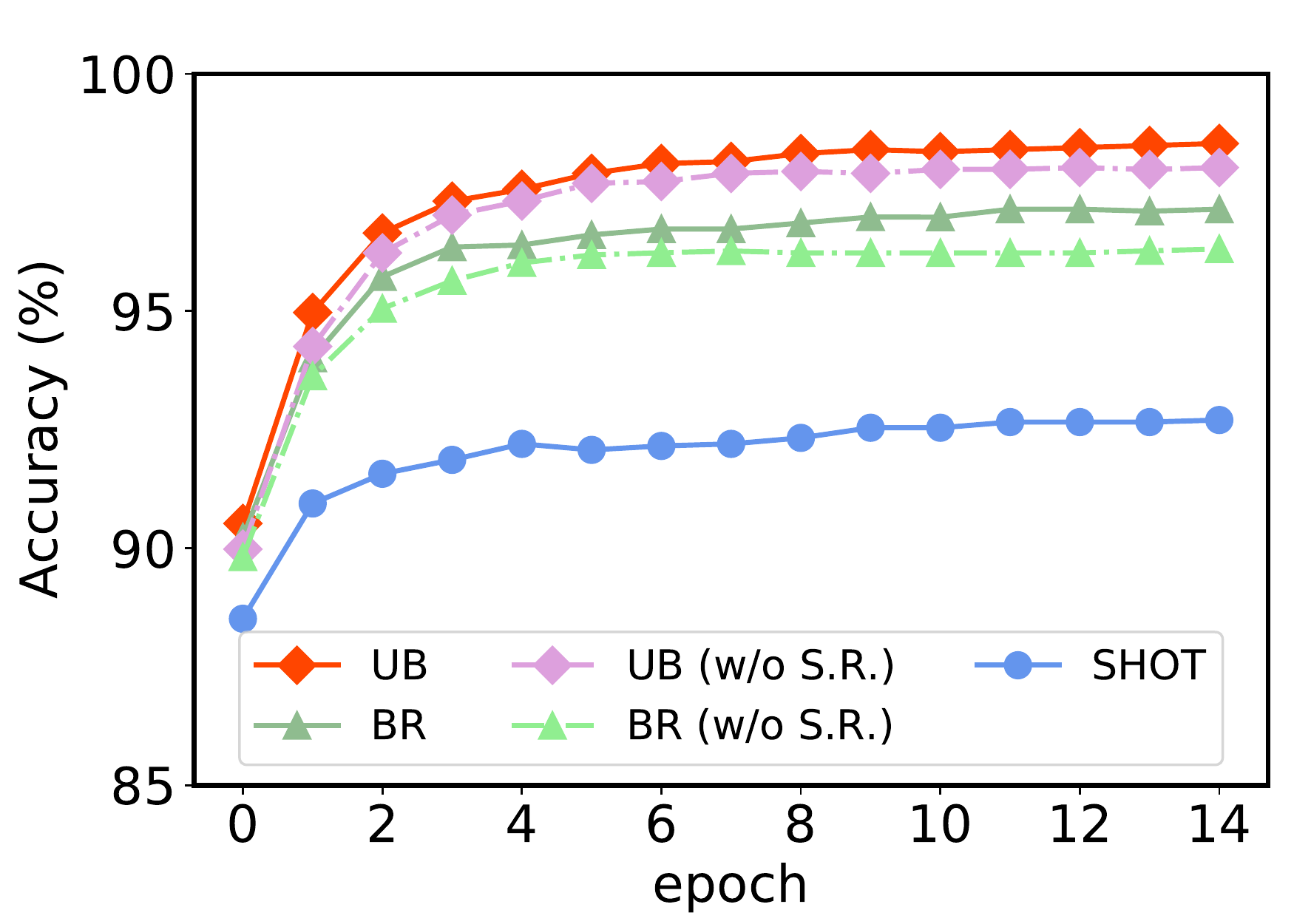} \label{fig:converge}}
		\end{minipage}
	\end{center}
	\setlength{\abovecaptionskip}{-0.05cm}
	\caption{(Left) Prior knowledge rectifies the pseudo label of ambiguous samples; (center) accuracies of pseudo labels before and after rectification using prior knowledge during the training of kSHOT on Office A$\rightarrow$W; (right) convergence curves on Office A$\rightarrow$W for comparison methods (S.R. is short for Smooth Regularization).}
	\label{fig:pk_guide}
\end{figure}

\begin{figure}[!t]
	\centering
	\includegraphics[width=0.16\textwidth]{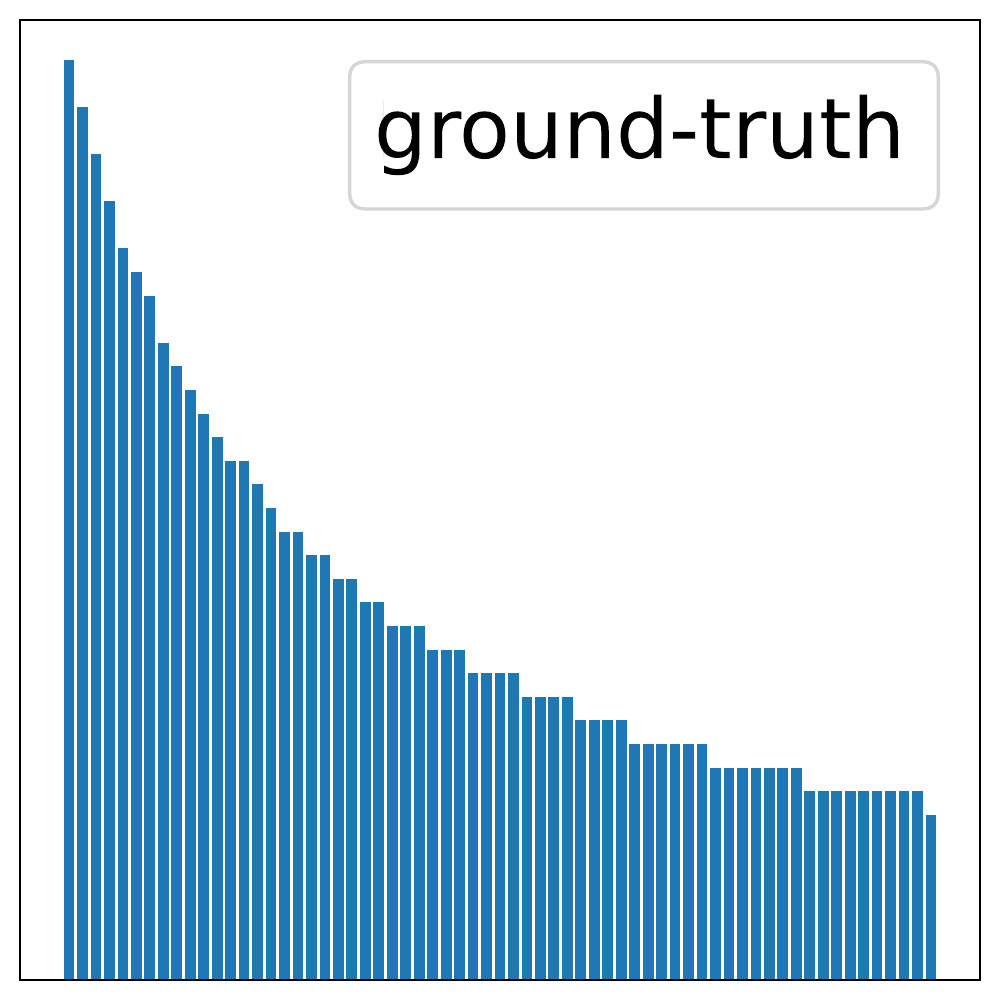}
	\includegraphics[width=0.16\textwidth]{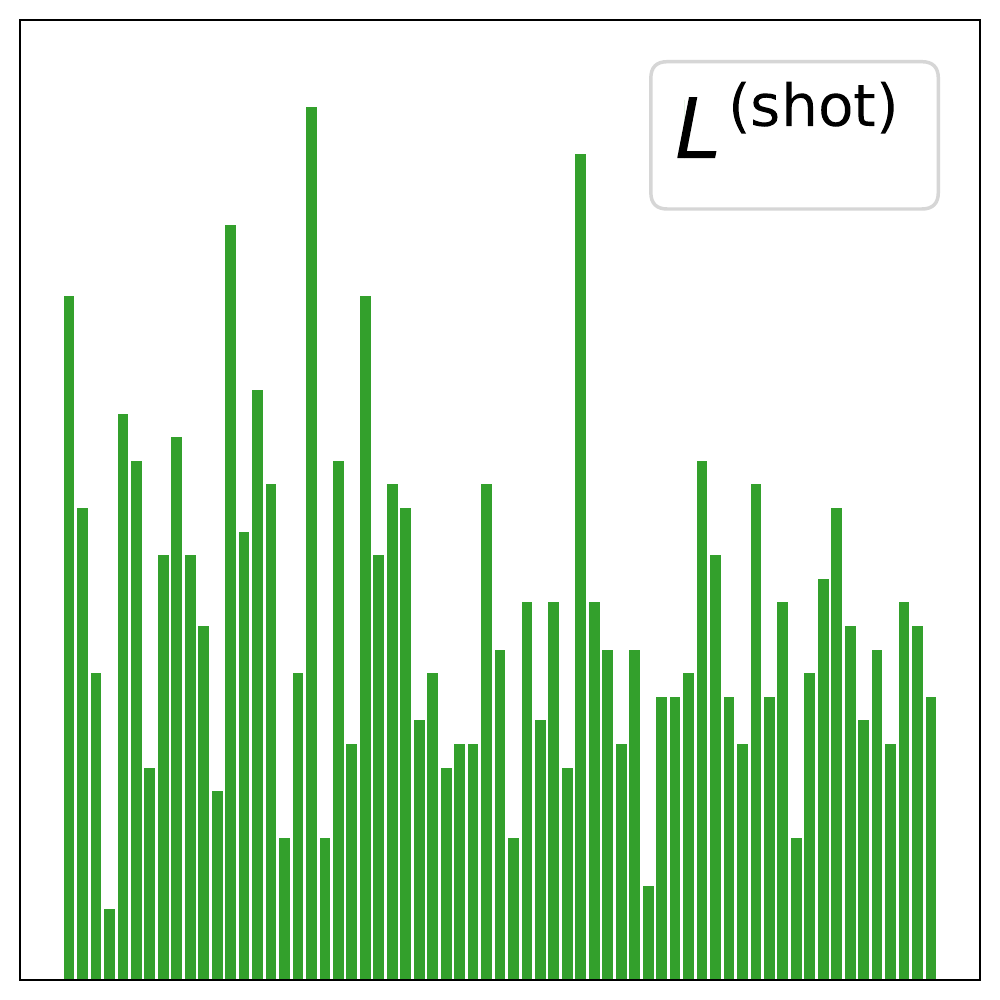}
    \includegraphics[width=0.16\textwidth]{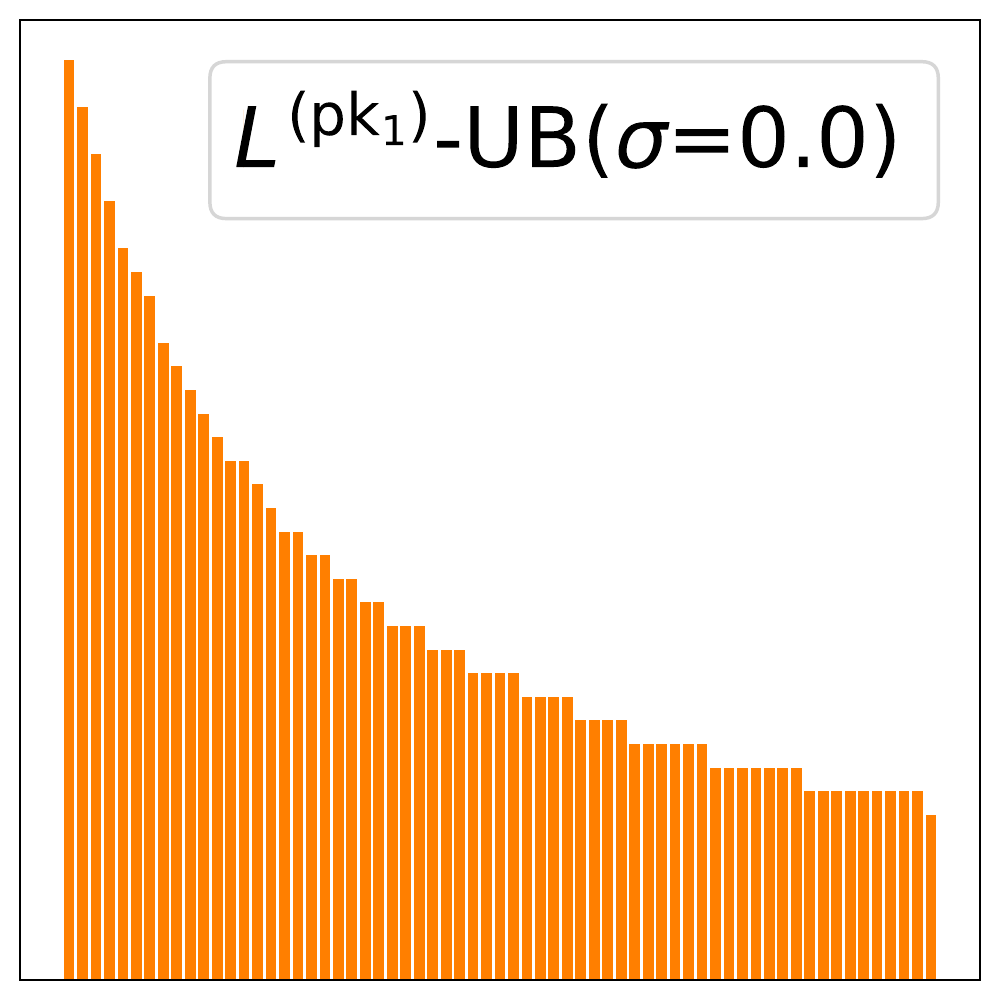}
    \includegraphics[width=0.16\textwidth]{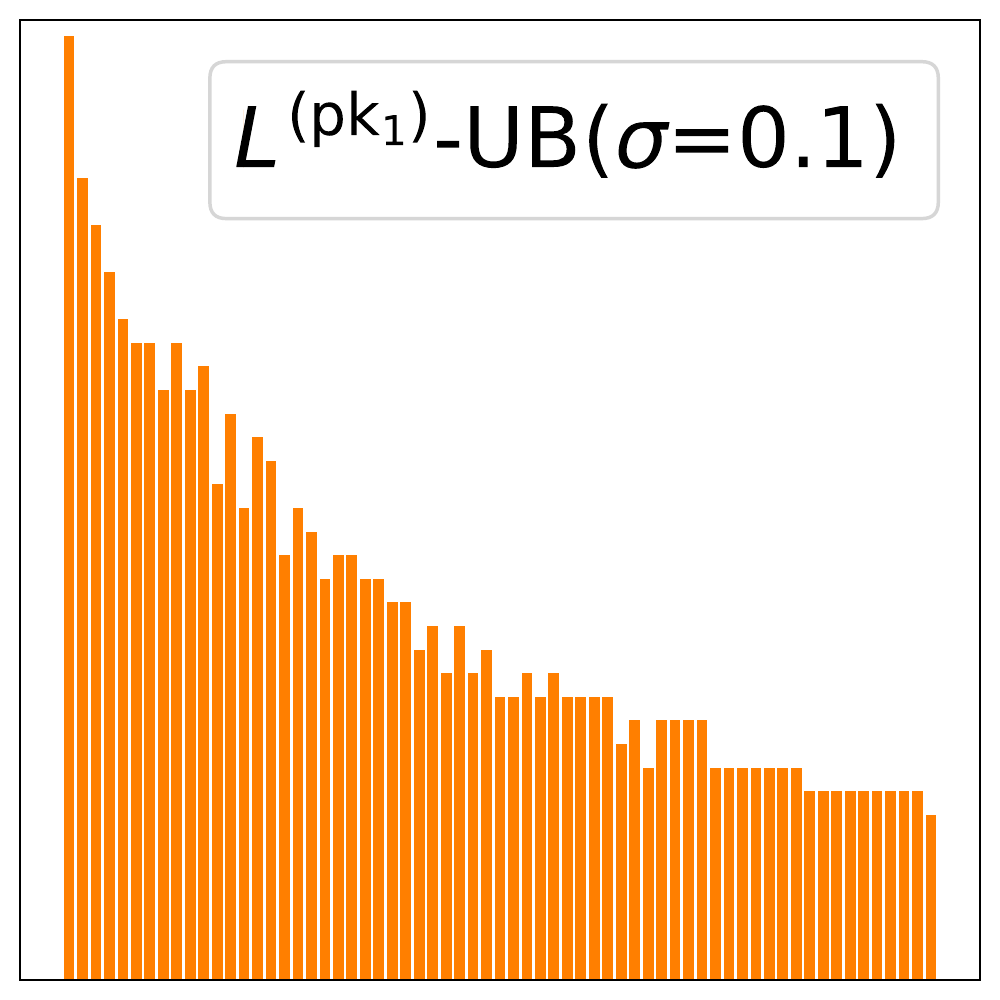}
    
    \includegraphics[width=0.16\textwidth]{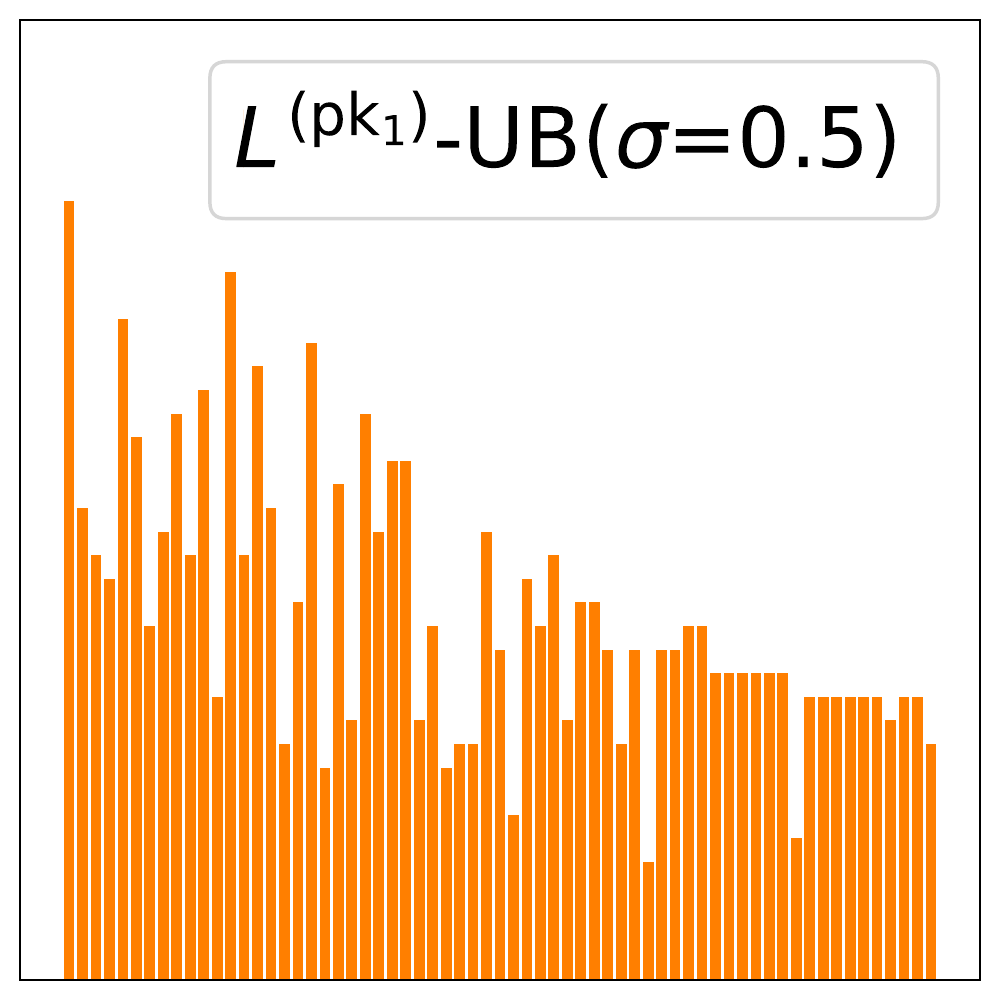}
    \includegraphics[width=0.16\textwidth]{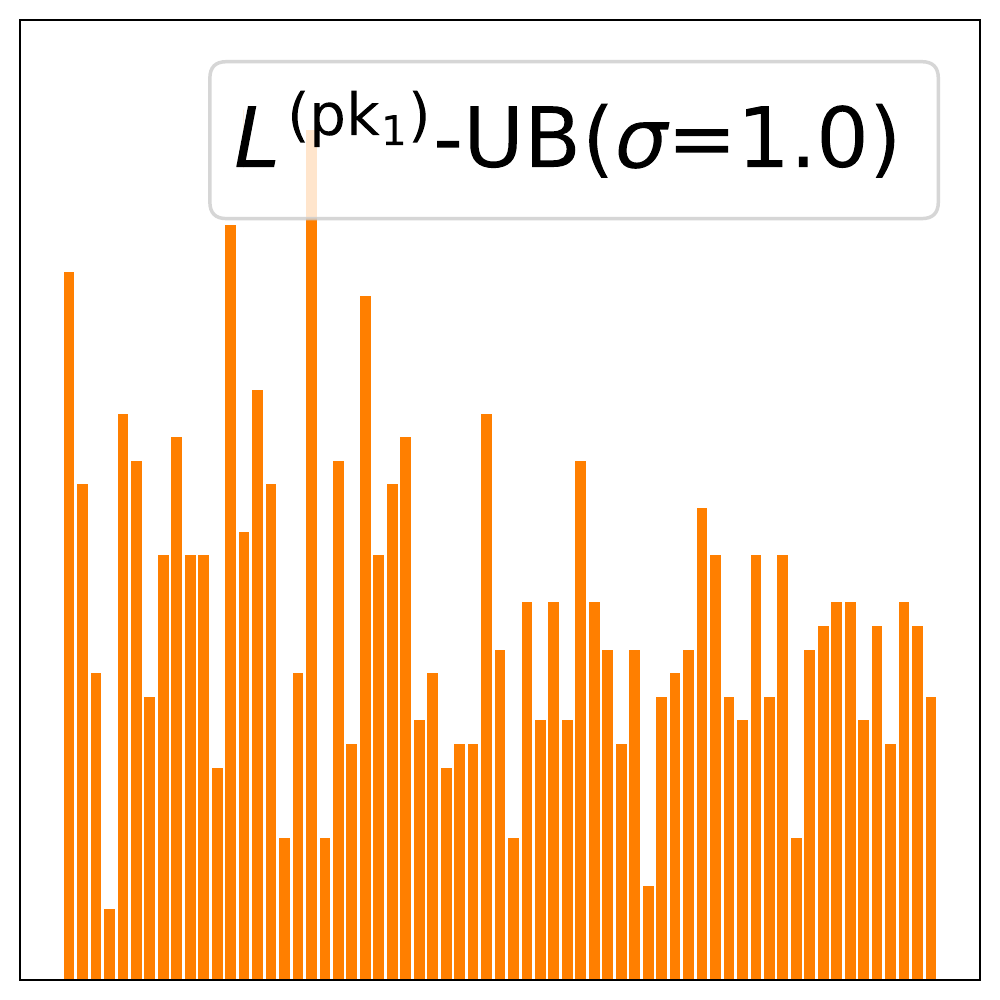}
    \includegraphics[width=0.16\textwidth]{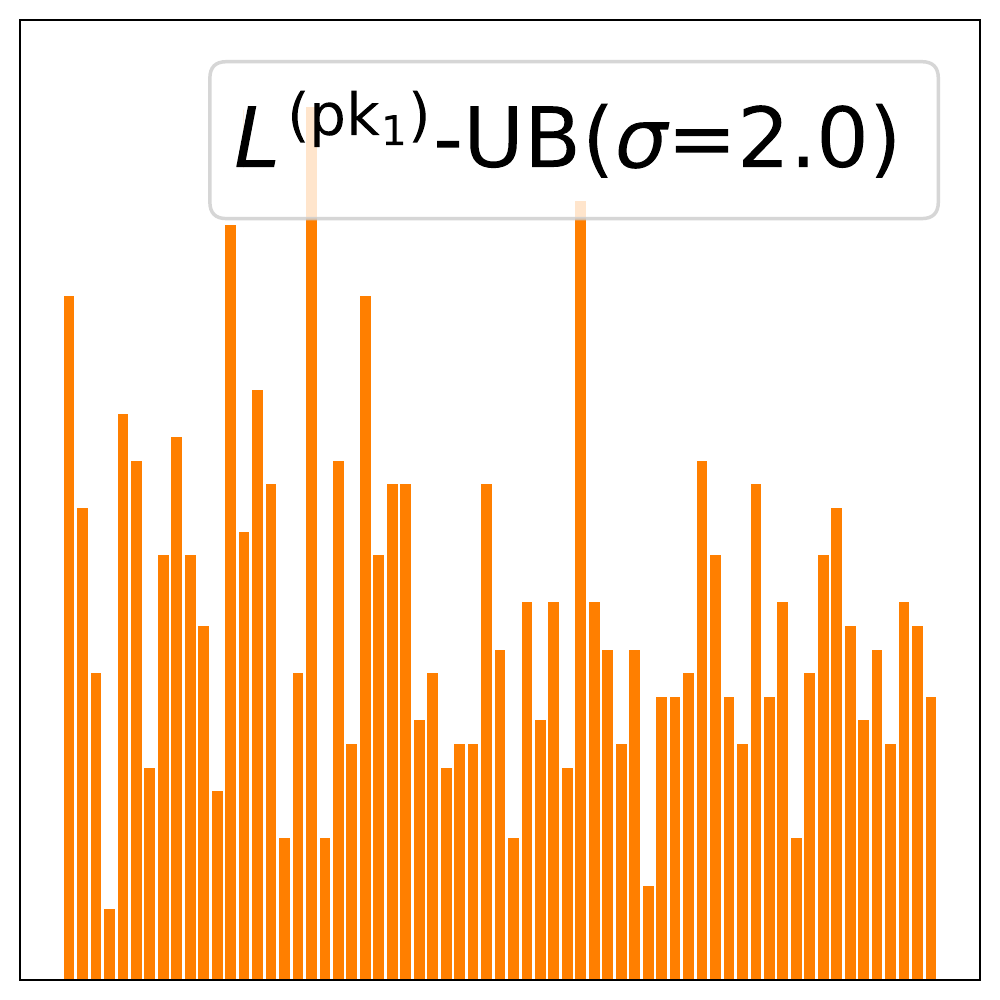}
    \includegraphics[width=0.16\textwidth]{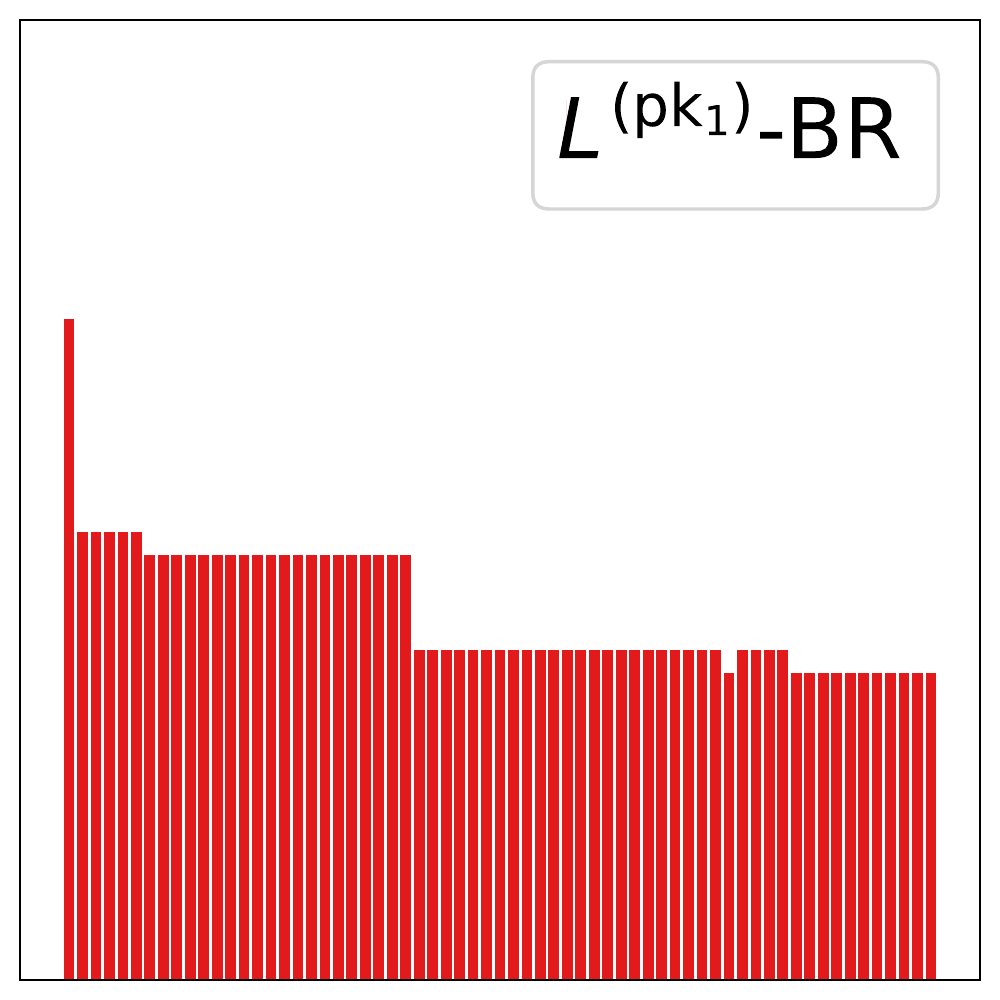}
	\caption{Prior knowledge rectifies distribution of pseudo labels in one pseudo-labeling step of kSHOT (Office-Home RS-UT P$\rightarrow$C).}
	\label{fig:hist}
\end{figure}

\begin{figure}[!t]
	\begin{center}
	\begin{minipage}{0.165\textwidth}
	\begin{flushright}
		\includegraphics[width=0.9\textwidth]{fig/hist/office-home-rsut_Product_RS_Clipart_UT_gt.pdf}
	\end{flushright}
	\end{minipage}
	\begin{minipage}{0.825\textwidth}
	\centering
	\includegraphics[width=0.18\textwidth]{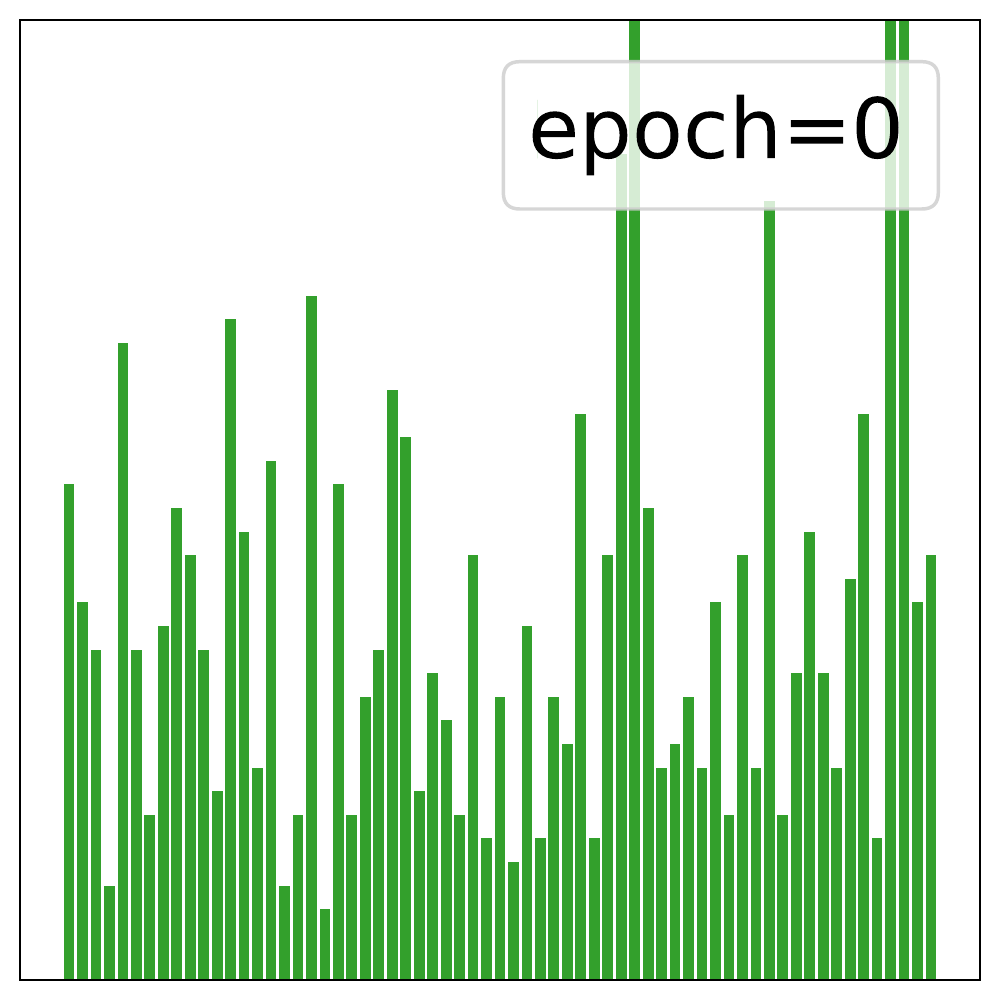}
	\includegraphics[width=0.18\textwidth]{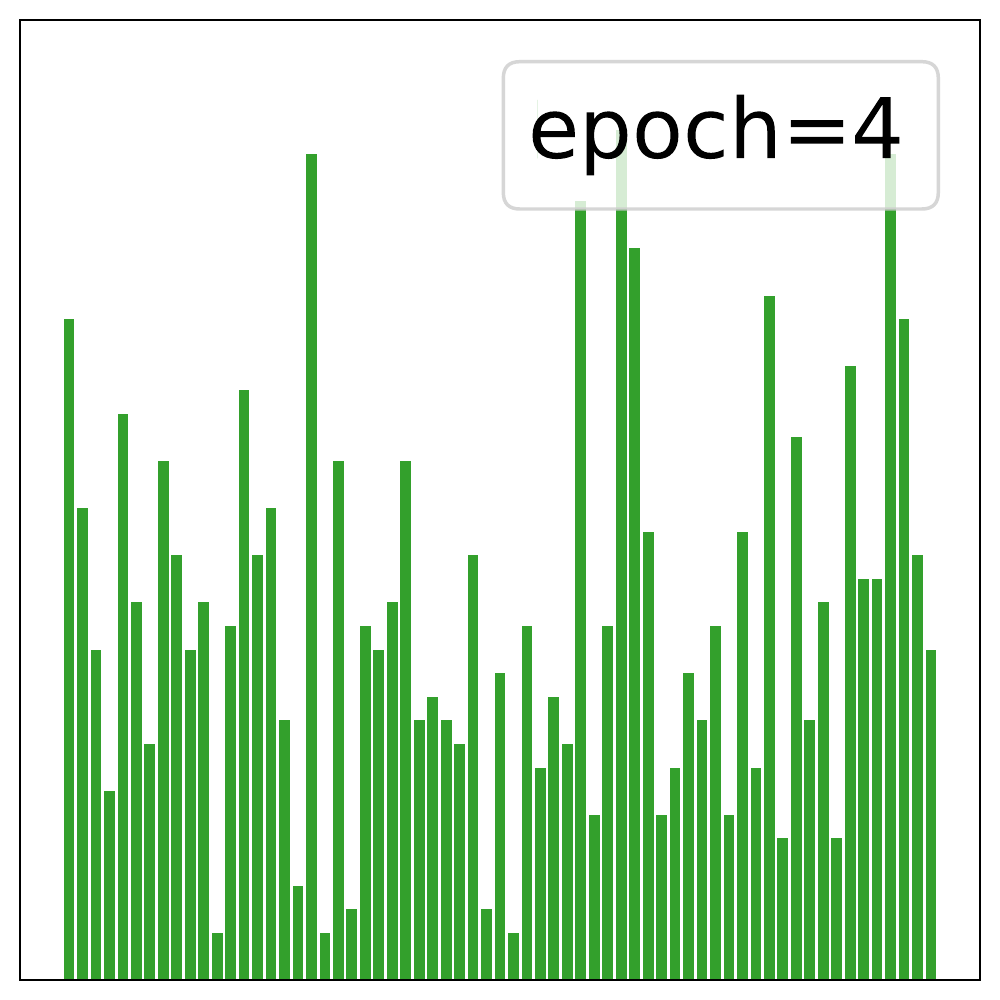}
	\includegraphics[width=0.18\textwidth]{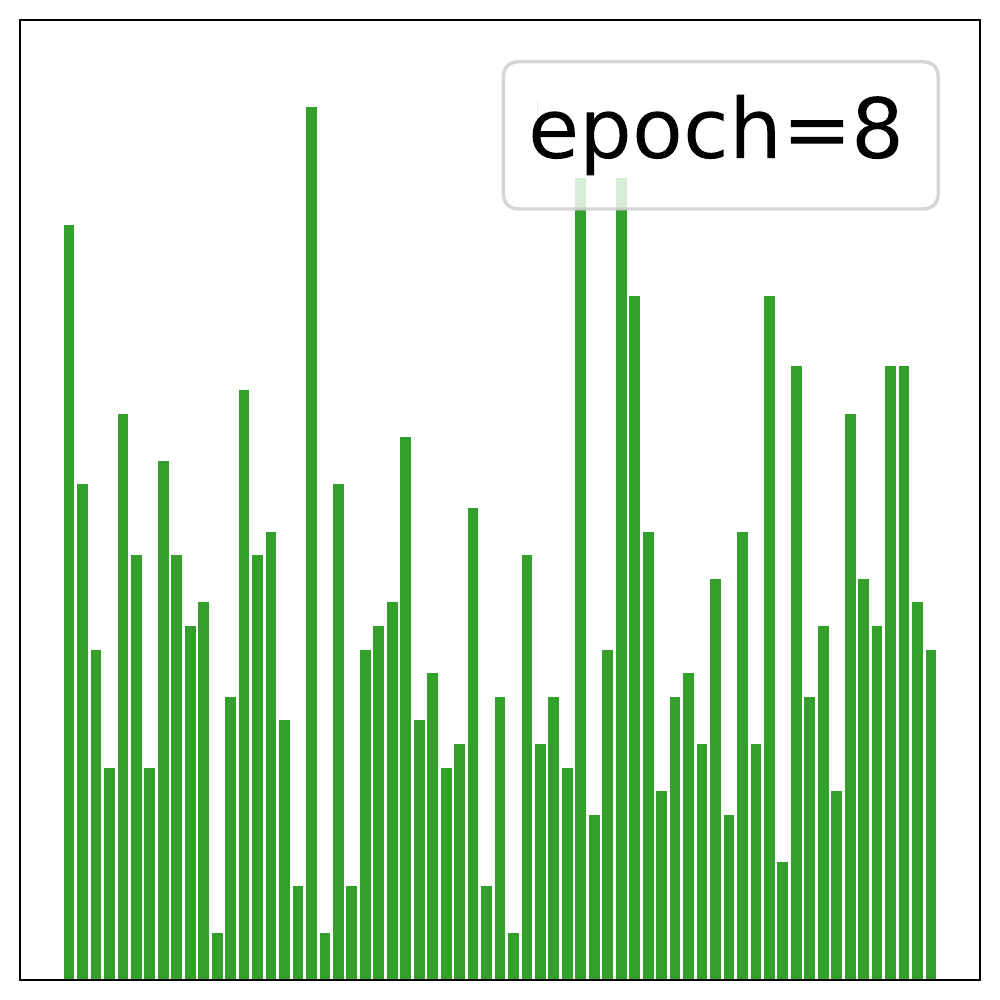}
	\includegraphics[width=0.18\textwidth]{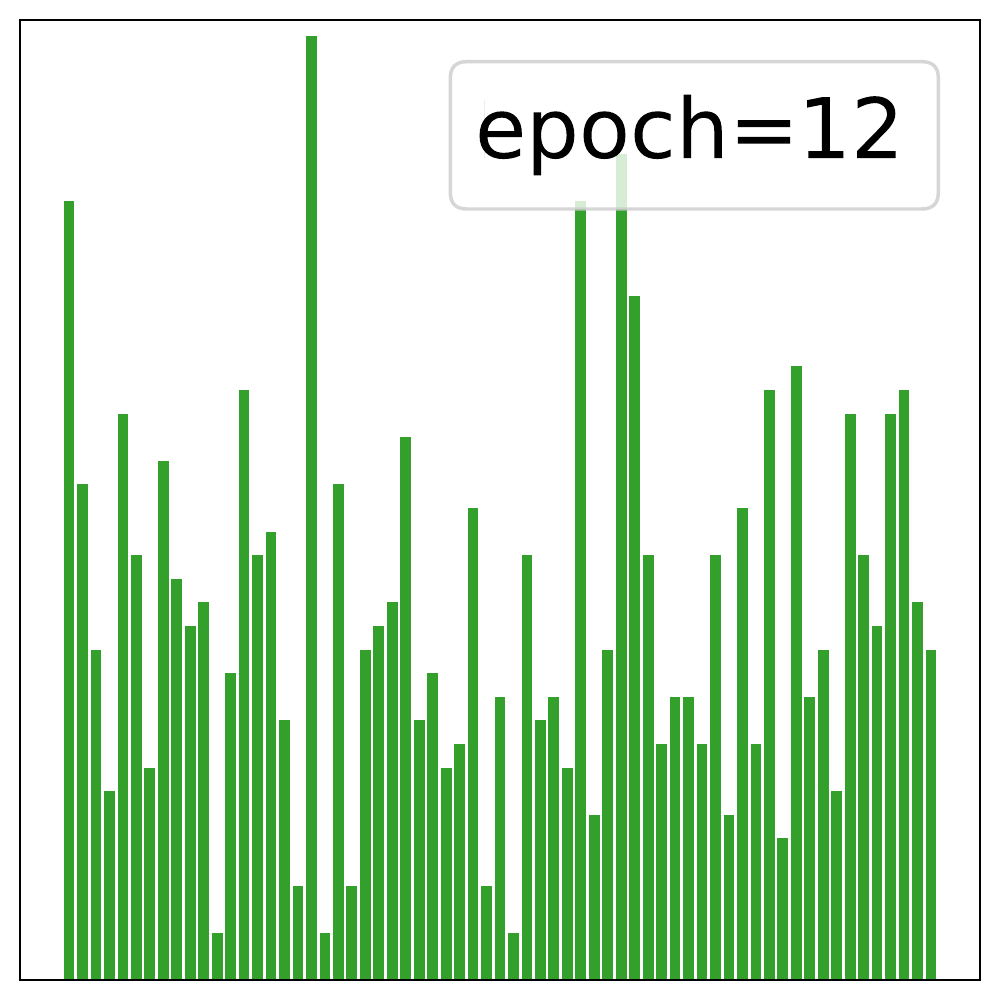}
	\includegraphics[width=0.18\textwidth]{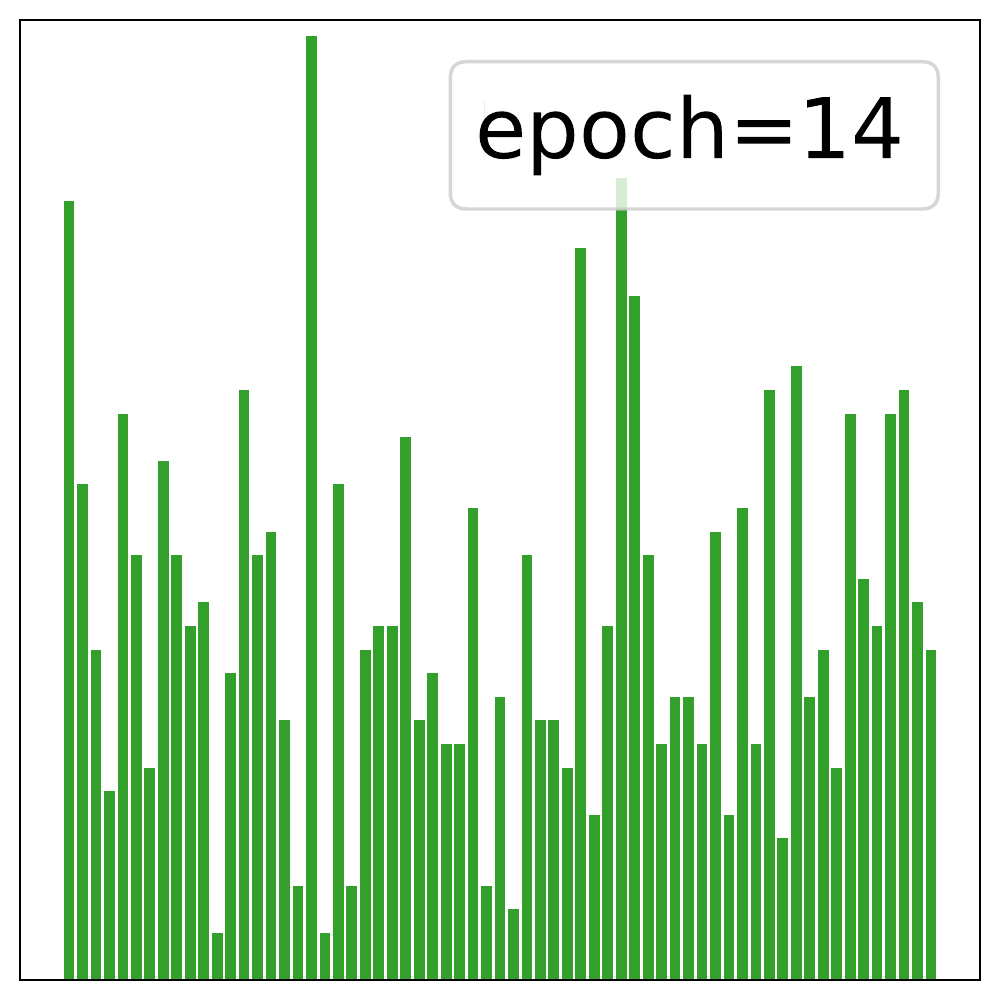}
	
	\includegraphics[width=0.18\textwidth]{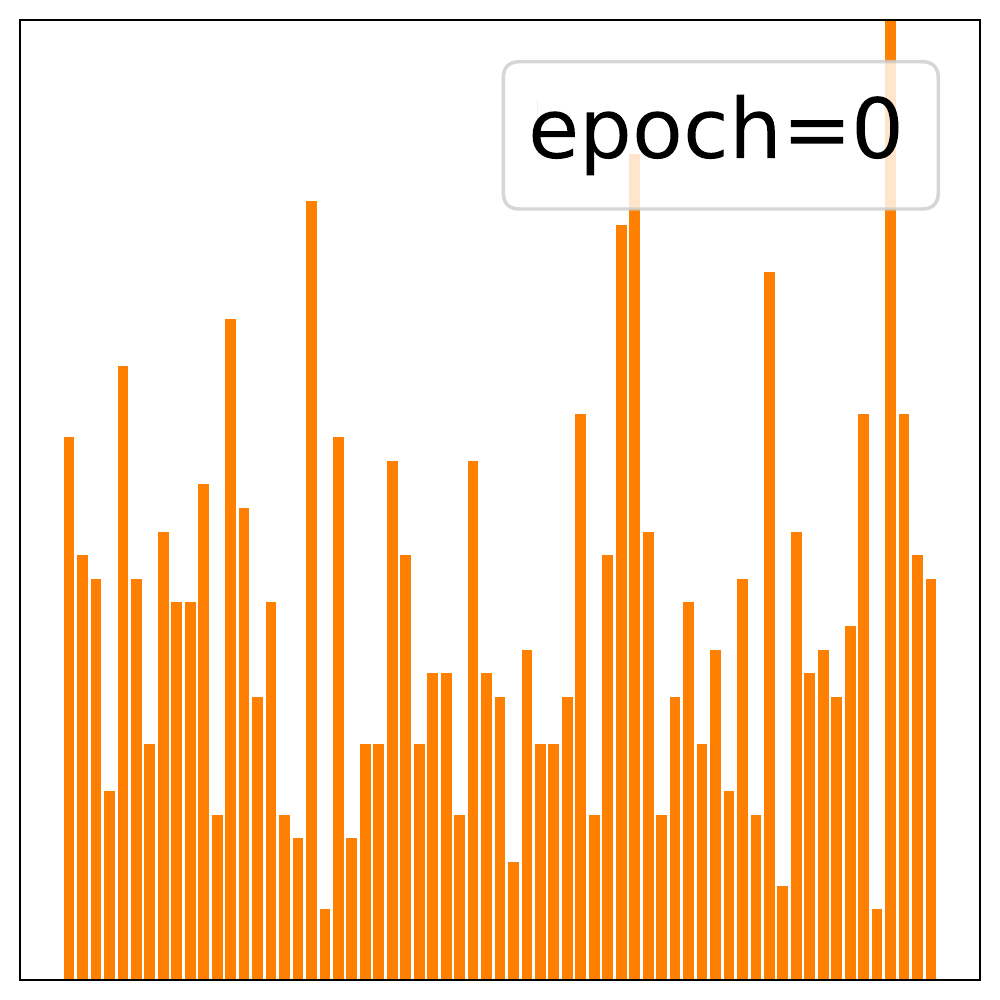}
	\includegraphics[width=0.18\textwidth]{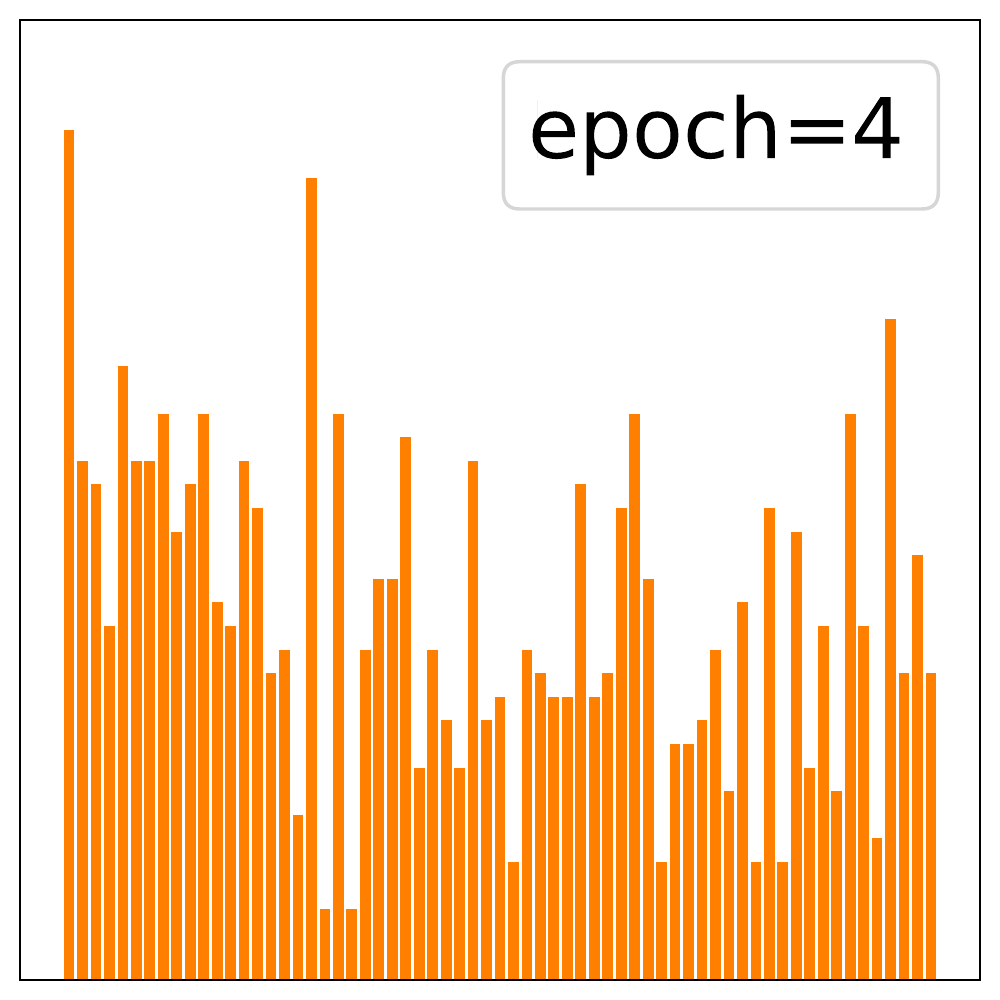}
	\includegraphics[width=0.18\textwidth]{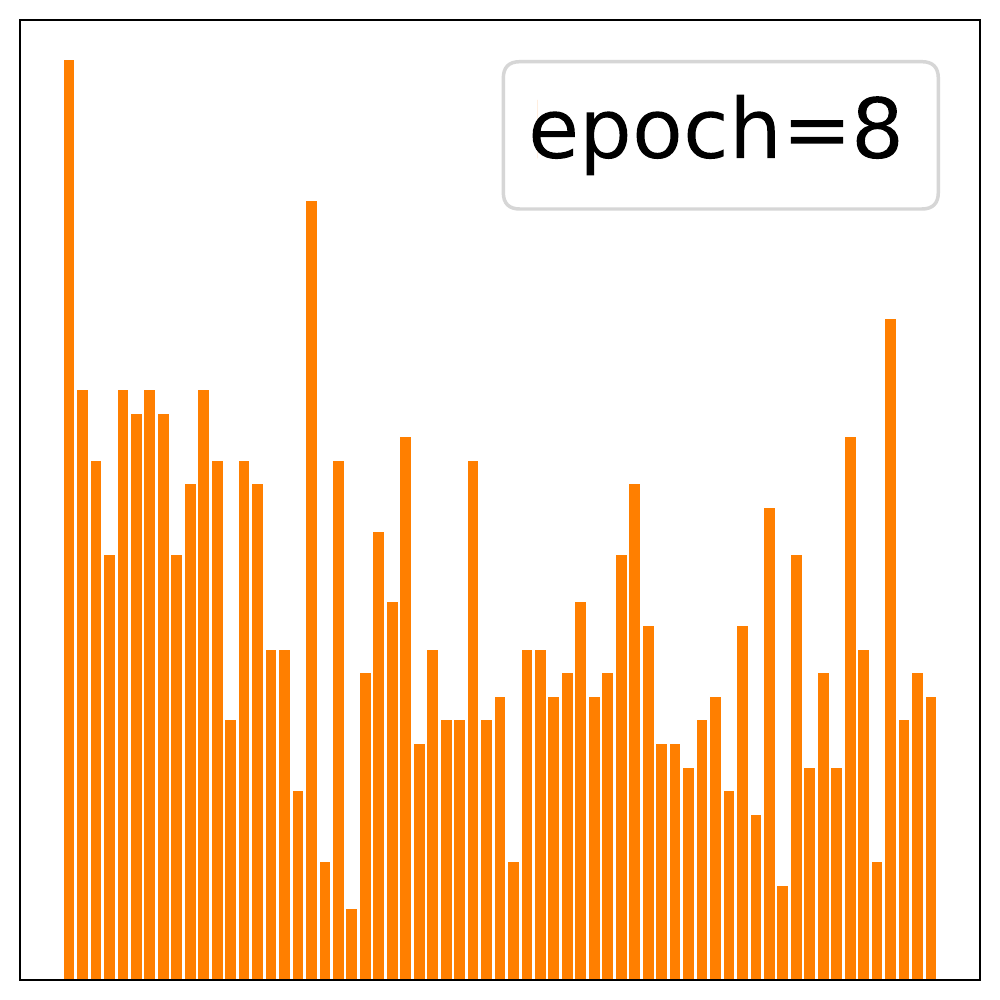}
	\includegraphics[width=0.18\textwidth]{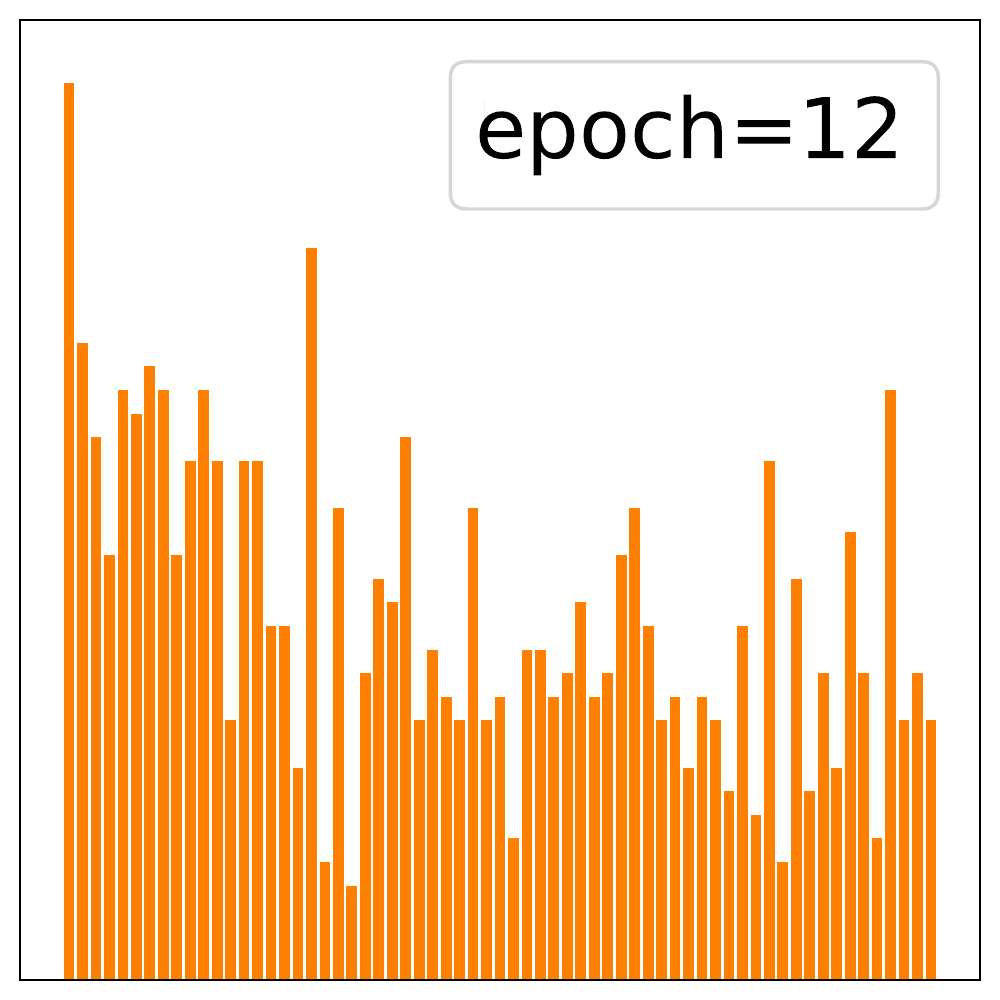}
	\includegraphics[width=0.18\textwidth]{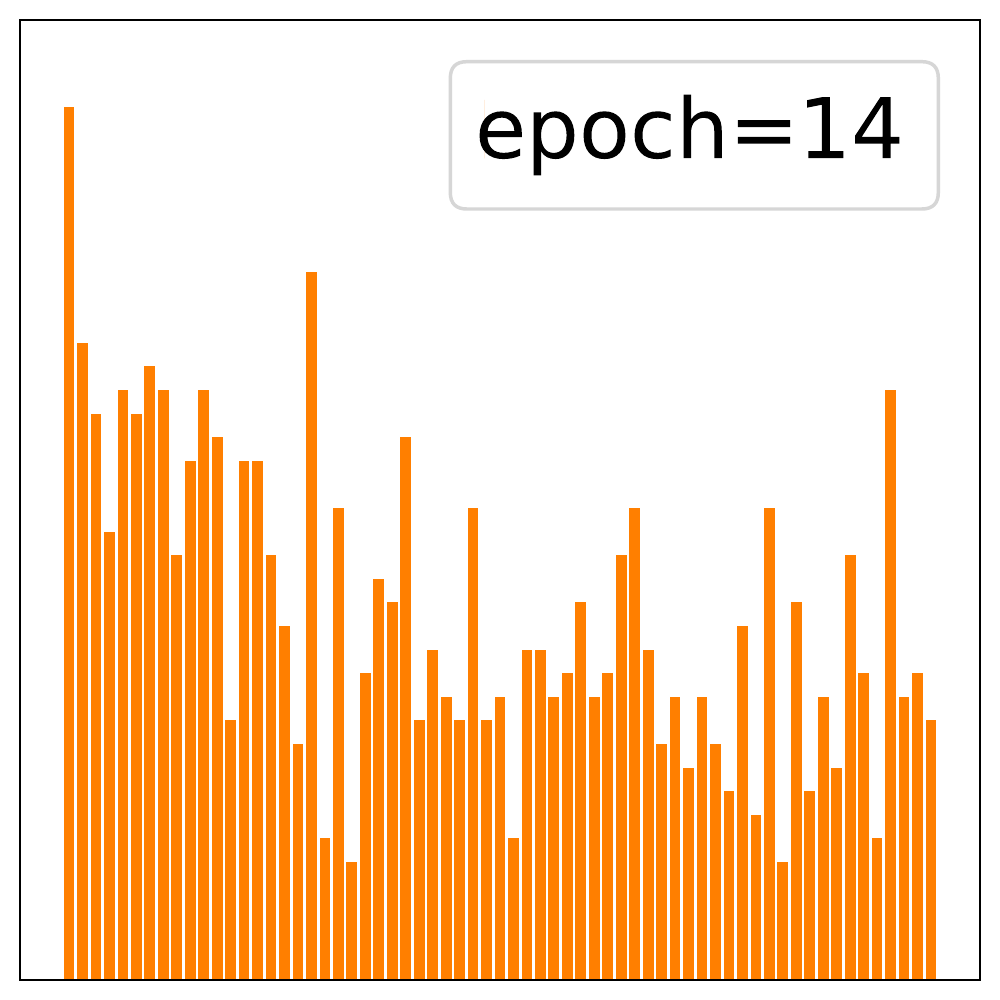}	
\end{minipage}	
\end{center}

\caption{Label distributions of network ($f_t$) predictions in SHOT (upper) and kSHOT (lower) (Office-Home RS-UT P$\rightarrow$C).}
\label{fig:hist_network}
\end{figure}

\textbf{How prior knowledge guides UDA?} To see why prior knowledge is helpful, we consider the following two aspects:
\begin{itemize}
	\item \textbf{Ambiguous samples}. As illustrated Fig.~\ref{fig:pk_guide} (left), our rectification module updates pseudo labels globally to match prior knowledge. This in effect moves decision boundaries in the feature space. The pseudo labels of ambiguous samples lying near the boundaries could be corrected during this process. Figure~\ref{fig:pk_guide} (center) plots the accuracies of three set of pseudo labels, $L^{({\rm shot})}$, $L^{(\rm pk_0)}$ and $L^{(\rm pk_1)}$, during the training process of kSHOT with UB($\sigma=0.0$) on Office A$\rightarrow$W. Clearly using prior knowledge obtains more accurate pseudo labels. This in turn benefits the subsequent self-training stage.

	\item \textbf{Label distribution}. Figure~\ref{fig:hist} plots distributions of ground-truth labels and pseudo labels on Office-Home RS-UT P$\rightarrow$C in one seudo-labeling step. The accuracy of pseudo labels is $\sim55\%$. As can be seen, the distribution of $L^{({\rm shot})}$ severely deviates from the ground-truth. In contrast, distributions of pseudo labels after rectification with prior knowledge are better compliant with the ground-truth. Figure~\ref{fig:hist_network} plots distributions of network predictions throughout the training process. The vanilla SHOT method hardly improves the label distribution after adaptation, whereas using prior knowledge drives it to be more similar to the ground-truth distribution in kSHOT.  
\end{itemize}

\begin{figure}[!t]
  \centering
  \includegraphics[width=0.32\linewidth]{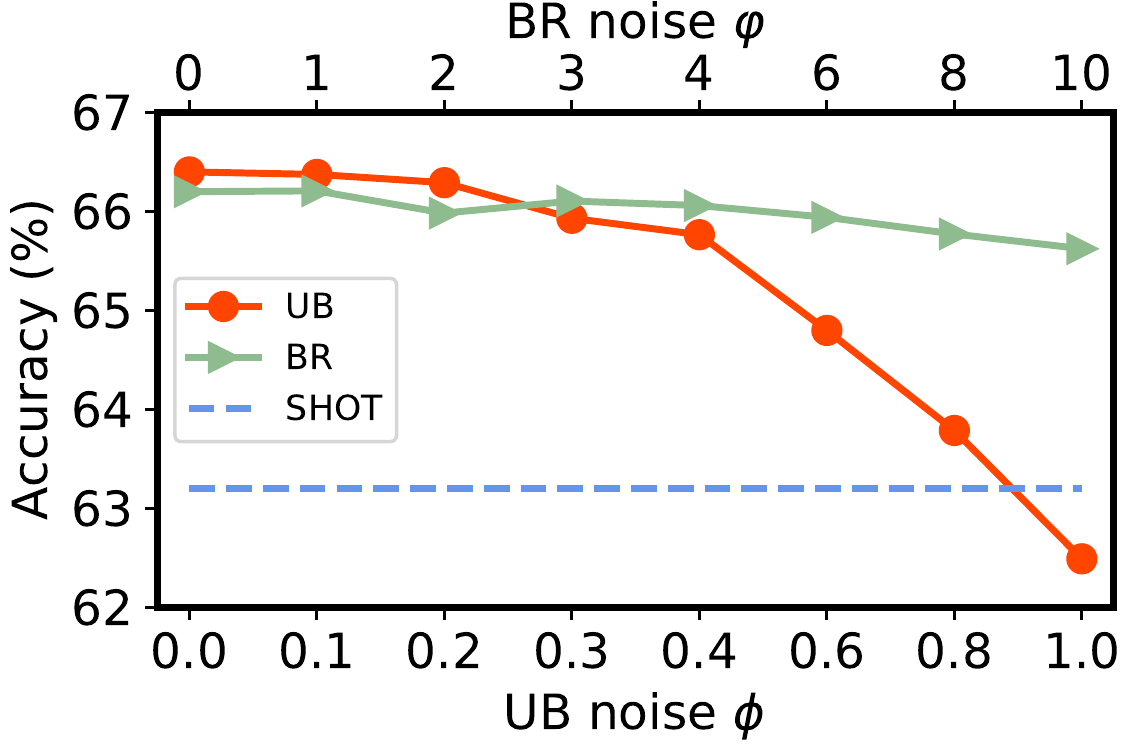}
  \includegraphics[width=0.32\linewidth]{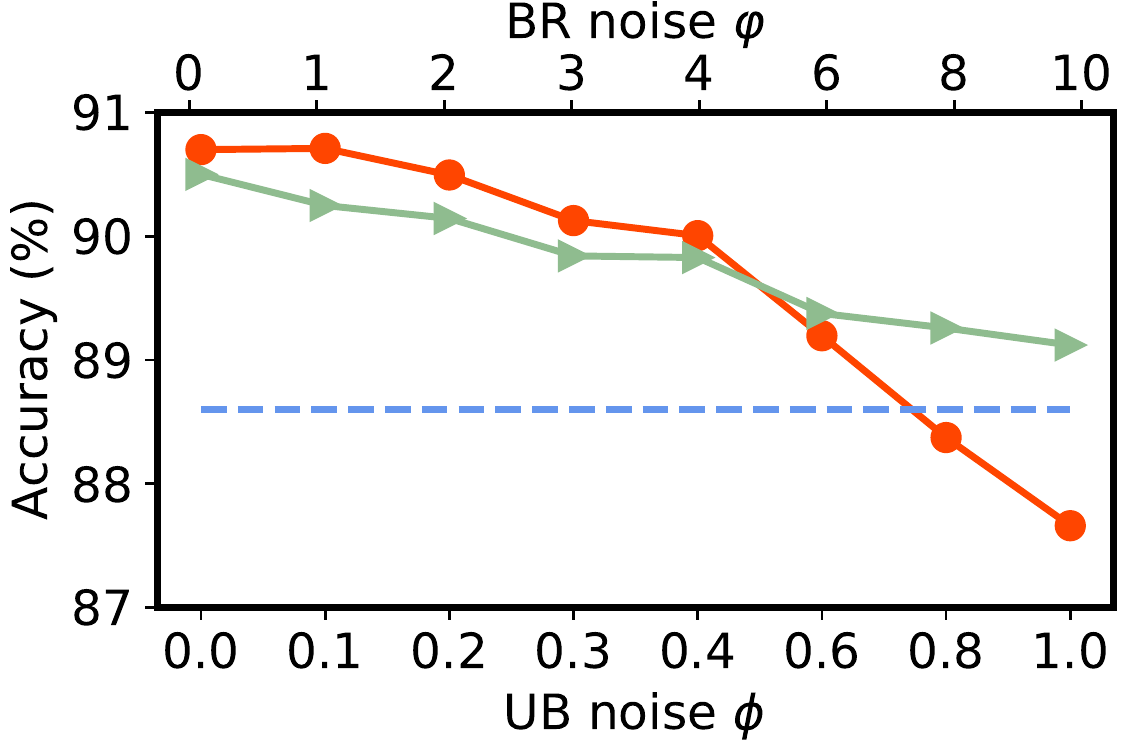}
   \includegraphics[width=0.32\linewidth]{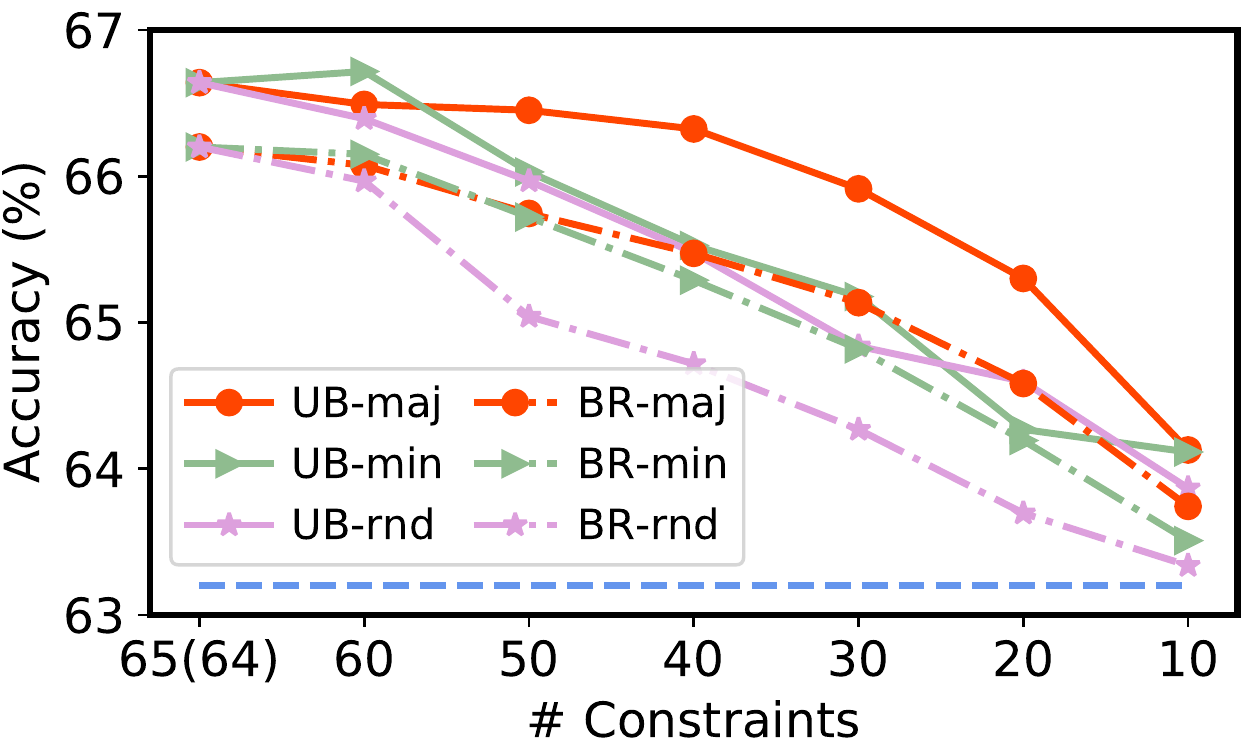}
   \caption{Using noisy prior knowledge in kSHOT on (left) Office-Home RS-UT and (center) Office; (right) using partial prior knowledge in kSHOT on Office-Home RS-UT.}
   \label{fig:pk_noise_NC}
\end{figure}

\noindent\textbf{Noisiness of the prior knowledge.} In practice, the prior knowledge might contain some level of noises. To study its effects, we manually add noises to the estimated class prior $q_c$. For UB, uniform noises are added through $\tilde{q}_c=q_c+\mathcal{U}(-q_c \phi, q_c \phi)$, where $\phi\in[0,1]$ controls the noise level. The noises have been centered to ensure that $\tilde{q}_c$ is a valid probability. Then $\tilde{q}_c$ is used to create the unary bound discussed in Sec.~\ref{sec:exp_setup}. For BR, after sorting all classes based on $q_c$, we randomly swap neighboring classes. Suppose the index of class $c_k$ in the sorted order is $I_{c_k}$, uniform noises are added through $\tilde{I}_{c_k}=I_{c_k}+\mathcal{U}(-\varphi,\varphi)$, where $\varphi$ controls the neighborhood size. Then we sort all classes based on the noisy indexes $\tilde{I}_{c_k}$, and use the resorted order to create binary relationship. Figure~\ref{fig:pk_noise_NC} (left and center) shows that under moderate noises, incorporating prior knowledge is still helpful and improves over the SHOT baseline.

\noindent\textbf{Completeness of the prior knowledge.} Until then, the prior constraints are assumed to cover every class. It is straightforward to generalize to partial constraints. We randomly select a portion of constraints that corresponds to the major (maj.), minor (min) or random (rnd) classes. Figure~\ref{fig:pk_noise_NC} (right) present the results under different number of selected constraints. Partial constrains imply less prior information, but still can benefit UDA training. Note that BR-rnd reduces the performance most, partially because the randomly selected binary relationship constraints hardly form the complete order of a subset of classes.

\noindent\textbf{Estimating class prior from partial data.} Table~\ref{tab:pk_subset} presents the experimental results when estimating $q_c$ from partial target data in kSHOT with UB($\sigma=0.0$) on VisDA-2017. As sampling ratio reduces, the estimation error (relative deviations from using full data) increases. Nevertheless, even when the average estimation error is about 15.5\% at a sampling ratio of 0.5\%, using prior knowledge still improves over SHOT by +1.16\% (82.9\% $\rightarrow$ 84.06\%).

\noindent\textbf{Effects of smooth regularization.} Table~\ref{tab:abl:sr} presents the ablation study on smooth regularization. Generally, using smooth regularization achieves comparable or better performance. The penalty depends on $|\mathcal{S}_t|=\sum_i \mathbb{I}[\bm{l}^{({\rm shot})}_i\neq\bm{l}^{(\rm pk_0)}_i]$, and varies across different tasks. In UB with large $\sigma$, the prior knowledge is not informative, hence $|\mathcal{S}_t|$ is small. We underline cases where using smooth regularization increases the accuracy significantly. The amount of improvement is most significant on tasks like R$\rightarrow$P, C$\rightarrow$R, C$\rightarrow$P. Comparing two types of prior knowledge, it is more helpful in BR. Since BR only tells the order of class probabilities, adding smooth regularization provides complementary information.

\begin{figtab}[!t]
\begin{minipage}{0.6\textwidth}
	\centering
	\setlength{\belowcaptionskip}{0.2cm}
	\setlength{\abovecaptionskip}{-0.05cm}
	\tabcaption{Ablating Smooth Regularization (S.R.) in kSHOT on {Office-Home RS-UT}. }
	\scalebox{0.88}{
	\begin{tabular}{p{0.3cm}p{0.5cm}<{\centering}p{0.6cm}<{\centering}|p{0.8cm}<{\centering}p{0.8cm}<{\centering}p{0.8cm}<{\centering}p{0.8cm}<{\centering}p{0.8cm}<{\centering}p{0.8cm}<{\centering}>{\columncolor{tbgray}}p{0.8cm}<{\centering}}
		\toprule
		& $\mathcal{K}$ & $\sigma$ & R$\shortrightarrow$P &	R$\shortrightarrow$C &	P$\shortrightarrow$R &	P$\shortrightarrow$C &	C$\shortrightarrow$R & C$\shortrightarrow$P & Avg.   \\ 	
		\midrule
		\multirow{6}{*}{\rotatebox{90}{wo/ S.R.}} & UB & 0.0 & 77.9 &	52.0 &	78.8 &	50.0 &	70.3 &	68.9 &	66.3 	\\
		& UB & 0.1 & 77.9 &	51.4 &	78.7 &	49.7 &	70.5 &	68.4 &	66.1 \\
		& UB & 0.5 & 76.6 &	50.5 &	77.0 &	48.4 &	67.7 &	67.3 &	64.6 \\
		& UB & 1.0 & 76.5 &	50.1 &	76.7 &	48.1 &	66.4 &	65.3 &	63.8  \\
		& UB & 2.0 & 76.4 &	49.9 &	75.8 &	47.8 &	65.1 &	64.2 &	63.2 \\
		& BR & -- & 77.9 &	51.2 &	78.1 &	49.2 &	69.0 &	67.6 &	65.5  \\
    	\midrule	
		\multirow{6}{*}{\rotatebox{90}{w/ S.R.}} & UB & 0.0 &\underline{78.8} &	51.3 &	\underline{79.1} &	49.8 &	\underline{71.3} &	\underline{69.7} &	{66.6} \\
		& UB & 0.1 & \underline{78.3} &	\underline{51.7} &	\underline{79.0} &	48.6 &	\underline{71.6} &	\underline{69.4} &	{66.4} \\
		& UB & 0.5 & {76.7} &	{50.6} &	\underline{77.3} &	48.4 &	\underline{68.4} &	67.3 &	{64.8} \\
		& UB & 1.0 & 76.3 &	50.0 &	{76.9} &	\underline{48.4} &	\underline{66.7} &	\underline{65.8} &	{64.0} \\
		& UB & 2.0 & 76.4 &	{50.0} &	{76.0} &	{47.9} &	{65.3} &	64.1 &	{63.3} \\
		& BR & -- & \underline{78.6} &	\underline{51.6} &	\underline{78.7} &	{49.3} &	\underline{70.1} &	\underline{68.8} &	{66.2} \\
		\bottomrule
	\end{tabular} }
	\label{tab:abl:sr}

\end{minipage}
\hfil
\begin{minipage}{0.36\textwidth}
\centering
\includegraphics[width=1.0\linewidth]{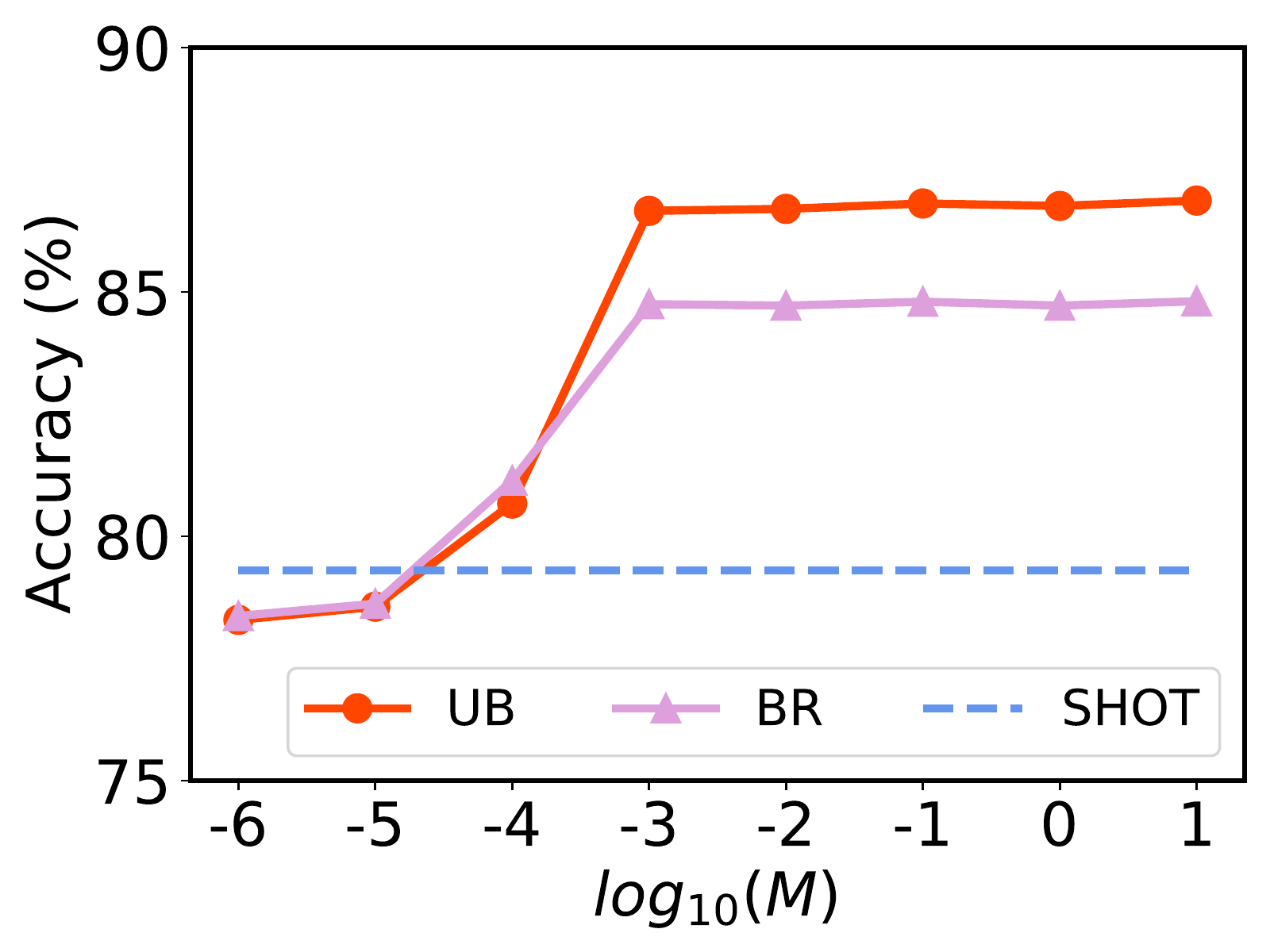}
	\figcaption{Effects of constant $M$ (by a scalar $n_t$) in kSHOT on Office-Home for PDA (averaged over 12 tasks).}
	\label{fig:M_pda}
\end{minipage}
\end{figtab}

\begin{table}[!t]
\caption{Estimating class prior from partial target data in kSHOT on VisDA-2017.} 
	\centering
	\begin{tabular}{p{3cm}p{1cm}<{\centering}p{1cm}<{\centering}p{1cm}<{\centering}p{1cm}<{\centering}p{1cm}<{\centering}p{1cm}<{\centering}}
	\toprule
	Sampling ratio & 0.5\% & 1\% & 5\% & 10\% & 50\% & 100\% \\
	\midrule
	Max. est. err (\%) & 47.5 & 25.2 & 11.2 & 9.6 & 4.1 & 0.0 \\
	Avg. est. err (\%) & 15.5 & 12.0 & 5.4 & 3.4 & 1.9 & 0.0 \\
	Avg. acc (\%) & 84.06 & 84.76 & 85.77 & 85.93 & 86.12 & 86.13\\
    \bottomrule
	\end{tabular} 
	\label{tab:pk_subset}
\end{table}

\noindent\textbf{Choice of constant $M$.} The prior knowledge is considered to be reliable, hence we expect the prior constraints to be satisfied in the rectified pseudo labels. To achieve this, $M$ need to be some large constant in Eq.~\ref{eq:pk_obj_scon_ub},\ref{eq:pk_obj_scon_br}. Figure~\ref{fig:M_pda} shows how $M$ affects the performance on Office-Home under PDA setting in kSHOT. When $M$ is very small, the method degenerates to SHOT. When $M$ is larger than some threshold (\eg, $10^{-3}\cdot n_t$), all soft constraints will in fact be satisfied. In our experiments, we use $M=10\cdot n_t$ for all tasks in both kSHOT and kDINE. 

\section{Conclusions}
We present a new yet realistic setting termed Knowledge-guided Unsupervised Domain Adaptation (KUDA). In KUDA, in addition to labeled source data and unlabeled target data, we have access to some prior knowledge about target label distribution. We present a novel rectification module that refines pseudo labels using prior knowledge through solving a constrained optimization problem. Then we integrate it into two representative self-training based methods, SHOT and DINE. Extensive experiments show that using prior knowledge can significantly improve the performance. We expect our work to inspire further investigations along the direction.

\clearpage

\appendix
\renewcommand{\thefigure}{A.\arabic{figure}}
\setcounter{figure}{0}
\renewcommand{\thetable}{A.\arabic{table}}
\setcounter{table}{0}
\renewcommand{\theequation}{A.\arabic{equation}}
\setcounter{equation}{0}

\section{Combining Two Types of Prior Knowledge}
In the paper, we considered two types of prior knowledge, \emph{Unary Bound} and \emph{Binary Relationship}. The two knowledge may have some overlapping. For example, it is possible to infer BR from UB when the unary bounds are tight, and vice versa. Nevertheless, when the bounds are not tight, one knowledge may provide complementary information for the other. Table~\ref{tab:supp:visda} lists results when combining UB and BR together in kSHOT on VisDA-2017. UB($\sigma=0.5$)+BR  performs slightly better than both UB($\sigma=0.5$) and BR. UB($\sigma=1.0$)+BR is on par with BR as UB($\sigma=1.0$) is not informative.

\begin{table}[!t]
	\centering
	\caption{Classification accuracies (\%) on \textbf{VisDA-2017}.}
	\scalebox{0.88}{
	\begin{tabular}{p{1.0cm}@{}p{1.6cm}<{\centering}@{}p{0.6cm}<{\centering}|@{}p{0.8cm}<{\centering}@{}p{0.8cm}<{\centering}@{}p{0.8cm}<{\centering}@{}p{0.8cm}<{\centering}@{}p{0.8cm}<{\centering}@{}p{0.8cm}<{\centering}@{}p{0.8cm}<{\centering}@{}p{0.8cm}<{\centering}@{}p{0.8cm}<{\centering}@{}p{0.8cm}<{\centering}@{}p{0.8cm}<{\centering}@{}p{0.8cm}<{\centering}@{}>{\columncolor{tbgray}[-0.2ex][0.35ex]}p{0.8cm}<{\centering}}
		\toprule
		Method & $\mathcal{K}$ & $\sigma$ & aero. & bike & bus & car & horse & knife & moto. & pers. & plant & sktb. & train & truck & Avg. \\ 	
		\midrule
		SHOT &-- & -- & 94.3 & 88.5 & 80.1 & 57.3 & 93.1 & 94.9 & 80.7 & 80.3 & 91.5 & 89.1 & 86.3 & 58.2 & 82.9 	\\
		\midrule
		\multirow{8}{*}{kSHOT} & UB & 0.0 & 95.7 &	88.7 &	\HL{81.4} &	\HL{73.4} &	94.7 &	94.2 &	\HL{88.1} &	82.5 &	93.4 &	91.1 &	87.2 &	\HL{63.1} & \HL{86.1} \\	
		& UB & 0.1 & 96.1 &	\HL{90.2} &	80.7 &	71.5 &	\HL{96.0} &	91.3 &	85.7 &	83.5 &	\HL{94.5} &	\HL{91.3} &	87.1 &	61.5 &  85.8 \\
		& UB & 0.5 & 95.2 &	89.6 &	79.7 &	59.6 &	94.8 &	90.7 &	82.0 &	\HL{86.2} &	92.7 &	90.2 &	86.8 &	59.8  &  83.9 \\
		& UB & 1.0 & 94.8 &	88.3 &	79.1 &	56.8 &	93.8 &	92.8 &	80.6 &	82.7 &	91.0 &	90.9 &	86.2 &	59.0  & 83.0  \\
		& UB & 2.0 & 94.6 &	87.7 &	78.9 &	55.9 &	93.4 &	94.8 &	80.2 &	81.4 &	89.3 &	89.9 &	86.1 &	58.6  & 82.6  \\	
		& BR &  -- & \HL{96.3} &	89.2 &	79.7 &	58.0 &	94.2 &	92.7 &	81.1 &	81.1 &	92.2 &	90.9 &	\HL{88.7} &	59.2  &  83.6\\
		& UB+BR & 0.5 & 95.8 &	89.1 &	81.1 &	60.2 &	95.1 &	91.5 &	84.3 &	82.7 &	93.4 &	91.4 &	88.7 &	59.8 &	84.4 \\
		& UB+BR & 1.0 & 96.2 &	89.1 &	79.7 &	58.0 &	94.2 &	92.6 &	81.1 &	81.1 &	92.2 &	90.9 &	88.7 &	59.3 &	83.6 \\
		\bottomrule
	\end{tabular} 
}
	\label{tab:supp:visda}
\end{table}

\begin{table}[!t]
	\caption{Classification accuracies (\%) on \textbf{Office-Home} for partial-set DA.}	
	\centering
	\scalebox{0.88}{
	\begin{tabular}{p{1.4cm}@{}p{0.6cm}<{\centering}@{}p{0.6cm}<{\centering}|@{}p{0.8cm}<{\centering}@{}p{0.8cm}<{\centering}@{}p{0.8cm}<{\centering}@{}p{0.8cm}<{\centering}@{}p{0.8cm}<{\centering}@{}p{0.8cm}<{\centering}@{}p{0.8cm}<{\centering}@{}p{0.8cm}<{\centering}@{}p{0.8cm}<{\centering}@{}p{0.8cm}<{\centering}@{}p{0.8cm}<{\centering}@{}p{0.8cm}<{\centering}@{}>{\columncolor{tbgray}[-0.2ex][0.35ex]}p{0.8cm}<{\centering}}
		\toprule
		Method & $\mathcal{K}$ & $\sigma$  & A$\shortrightarrow$C & A$\shortrightarrow$P & A$\shortrightarrow$R & C$\shortrightarrow$A & C$\shortrightarrow$P & C$\shortrightarrow$R &  P$\shortrightarrow$A & P$\shortrightarrow$C & P$\shortrightarrow$R & R$\shortrightarrow$A & R$\shortrightarrow$C & R$\shortrightarrow$P & Avg.   \\ 	
		\midrule
		SHOT & -- & -- & 64.8 & 85.2 & 92.7 & 76.3 & 77.6 & 88.8 & 79.7 & 64.3 & 89.5 & 80.6 & 66.4 & 85.8 & 79.3\\
		\midrule
		\multirow{2}{*}{kSHOT} & UB & 0.0 & \HL{74.1} &	\HL{94.4} &	\HL{94.3} &	\HL{84.3} &	\HL{93.1} &	\HL{93.0} &	\HL{85.3} &	\HL{73.4} &	\HL{93.5} &	\HL{86.7} &	\HL{74.7} &	\HL{95.0} &	\HL{86.8} \\
		& BR & -- & 72.2 &	92.9 &	92.8 &	82.3 &	89.8 &	90.9 &	83.6 &	69.6 &	92.6 &	86.0 &	71.7 &	93.3 &	84.8 \\			
			\midrule
			DINE & -- & -- & 58.1 & 83.4 & 89.2 & 78.0 & 80.0 & 80.6 & 74.2 & 56.6 & 85.9 & 80.6 & 62.9 & 84.8 & 76.2\\
			DINE$^*$ & -- & --  & 54.9 &	80.8 &	87.3 &	70.3 &	75.2 &	78.8 &	70.9 &	51.2 &	85.7 &	78.1 &	58.3 &	84.1 &	73.0  \\
			\midrule
			\multirow{2}{*}{kDINE} & UB & 0.0 & \HL{65.5}	 & \HL{91.4} &	\HL{92.3} &	\HL{80.2} &	\HL{89.3} &	\HL{91.2} &	\HL{81.6} &	\HL{64.4} &	\HL{91.6} &	\HL{84.5} &	\HL{69.3} &	\HL{93.2} &	\HL{82.9}  \\
			& BR & -- & 62.5 &	89.2 &	91.1 &	77.3 &	85.0 &	87.2 &	78.5 &	60.3 &	90.3 &	83.4 &	67.1 &	90.3 &	80.2 \\		
			\bottomrule
		\end{tabular} 
	}
	\label{tab:supp:officehome:pda}
\end{table}

\section{Visualization of Standard Deviations}
\begin{figure}[h]
	\centering
	\includegraphics[width=0.95\textwidth]{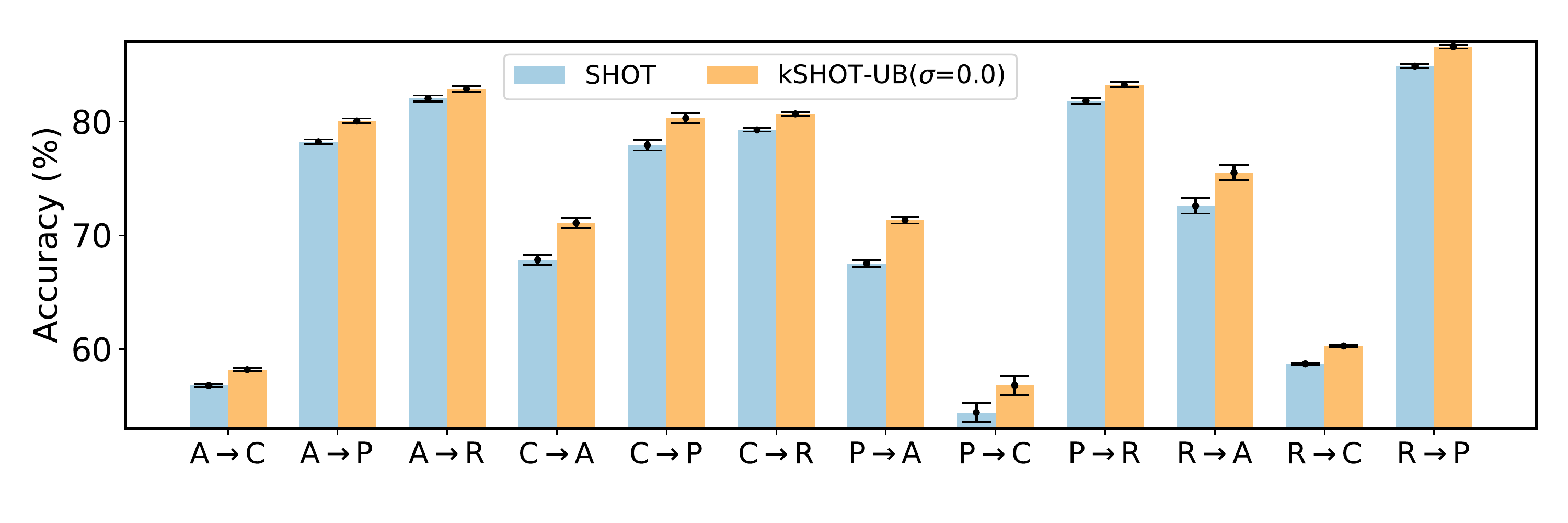}
	\includegraphics[width=0.95\textwidth]{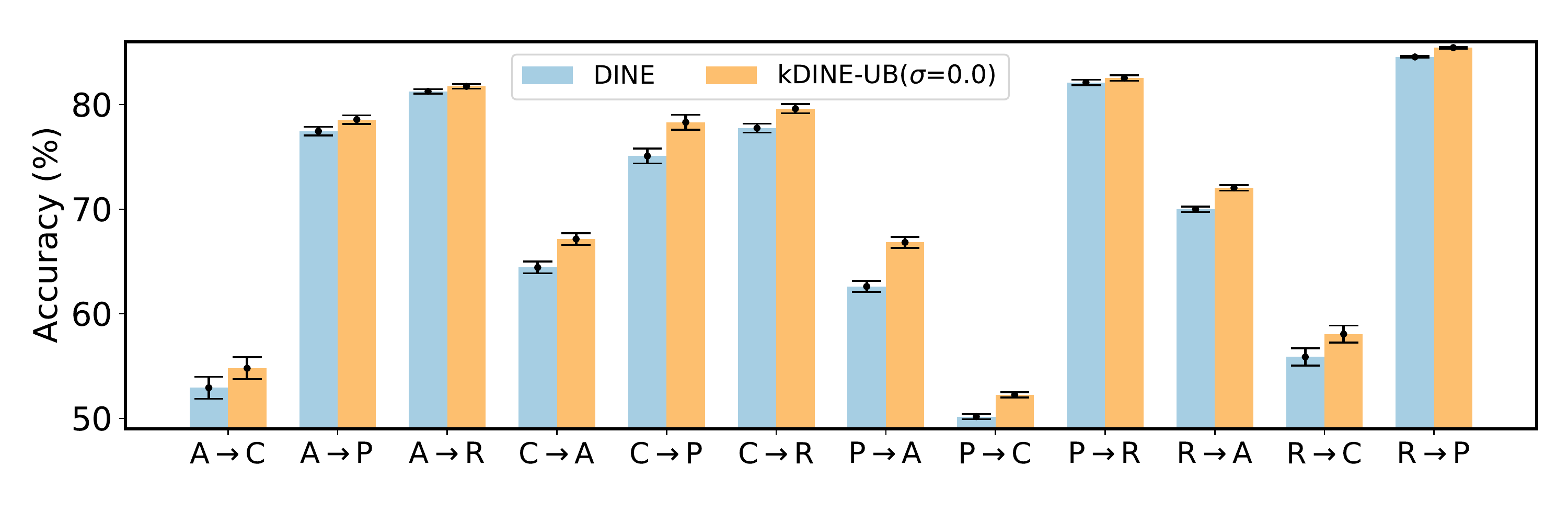}
	\caption{Visualization of standard deviations on Office-Home.}
	\label{fig:supp:std}
\end{figure}

For all tables and figures in the paper, we report the mean evaluation results of three repeated experiments with different random seeds. Figure~\ref{fig:supp:std} visualizes the standard deviations on Office-Home of comparison methods. As can be seen, the performances are stable to different initializations.

\section{More Detailed Results}
Table~\ref{tab:supp:officehome:pda} presents the detailed results on Office-Home for partial-set DA. Accuracies per class on VisDA-2017 are listed in Tab.~\ref{tab:supp:visda}.

\section{Effects of Label Smoothing in kDINE}
DINE~\cite{liang2021dine} distillates knowledge from the source predictor to a target model. As mentioned in~\cite{liang2021dine}, using label smoothing for the teacher probability is superior to using one-hot encoding. To show its effect in kDINE, we compare a variant of kDINE without label smoothing. The objective function is
\begin{equation}\label{eq:supp:dine_obj_pk}
	\mathcal{L}_{\rm kdine}^*= \mathbb{E}_{\bm{x}^t_i}\mathcal{D}_{\rm kl}\left(\frac{P^{\rm tch}(\bm{x}^t_i)+ \bm{l}_i^{(\rm pk_1)}}{2}  \middle\| f_t(\bm{x}^t_i)\right) + \beta \mathcal{L}_{\rm mix} - \mathcal{L}_{\rm im}
\end{equation} 
Compared with Eq.~\ref{eq:dine_obj_pk} of the paper, the smoothed label $\tilde{\bm{l}}_i^{(\rm pk_1)}$ is replaced with the one-hot label $\bm{l}_i^{(\rm pk_1)}$. We term this variant as kDINE$^*$. For a fair comparison with DINE, we also replace $\bm{l}_i^{(\rm pk_1)}$ with $\bm{l}_{i}^{({\rm dine})}$ in Eq.~\ref{eq:supp:dine_obj_pk} and term it as DINE$^{**}$. Table~\ref{tab:supp:officehome_dine} shows that using one-hot labels indeed degrades the performance in both DINE$^{**}$ and kDINE$^*$. Nevertheless, with class prior knowledge, kDINE$^*$ still achieves much better accuracies than DINE$^{**}$. This verifies the effectiveness of considering prior knowledge and  our proposed rectification module.

\begin{table}[!t]
	\caption{Classification accuracies (\%) on \textbf{Office-Home}.}	
	\centering
	\scalebox{0.88}{
	\begin{tabular}{p{1.4cm}@{}p{0.6cm}<{\centering}@{}p{0.6cm}<{\centering}|@{}p{0.8cm}<{\centering}@{}p{0.8cm}<{\centering}@{}p{0.8cm}<{\centering}@{}p{0.8cm}<{\centering}@{}p{0.8cm}<{\centering}@{}p{0.8cm}<{\centering}@{}p{0.8cm}<{\centering}@{}p{0.8cm}<{\centering}@{}p{0.8cm}<{\centering}@{}p{0.8cm}<{\centering}@{}p{0.8cm}<{\centering}@{}p{0.8cm}<{\centering}@{}>{\columncolor{tbgray}[-0.2ex][0.35ex]}p{0.8cm}<{\centering}}
			\toprule
			Method & $\mathcal{K}$ & $\sigma$ & A$\shortrightarrow$C & A$\shortrightarrow$P & A$\shortrightarrow$R & C$\shortrightarrow$A & C$\shortrightarrow$P & C$\shortrightarrow$R &  P$\shortrightarrow$A & P$\shortrightarrow$C & P$\shortrightarrow$R & R$\shortrightarrow$A & R$\shortrightarrow$C & R$\shortrightarrow$P & Avg.   \\ 	
			\midrule
			DINE & -- & -- & 52.2 & 78.4 & 81.3 & 65.3 & 76.6 & 78.7 & 62.7 & 49.6 & 82.2 & 69.8 & 55.8 & 84.2 & 69.7	\\
			DINE$^{*}$ & -- & --  & 51.8	 &76.0 &	79.6 &	63.1 &	75.1 &	76.5 &	60.4 &	48.5 &	80.7 &	69.4 &	55.9 &	83.5 &	68.4  \\
			DINE$^{**}$ & -- & --  & 51.3 &	75.4 &	79.2 &	62.7 &	74.6 &	75.8 &	59.8 &	48.1 &	80.2 &	68.9 &	55.7 &	83.1 &	67.9 \\
			\midrule
			\multirow{6}{*}{kDINE} & UB & 0.0 & 54.8 &	78.6 &	\HL{81.7} &	\HL{67.1} &	78.3 &	\HL{79.6} &	\HL{66.8} &	\HL{52.3} &	82.5 &	\HL{72.0} &	\HL{58.1} &	\HL{85.4} &	\HL{71.4} \\	
			& UB & 0.1 & \HL{55.0} &	78.8 &	81.1 &	66.4 &	77.7 &	79.2 &	66.4 &	51.8 &	82.3 &	71.5 &	58.0 &	84.9 &	71.1 \\
			& UB & 0.5 & 52.9 &	76.7 &	79.9 &	64.5 &	76.3 &	77.8 &	63.8 &	51.0 &	80.9 &	70.5 &	57.1 &	84.2 &	69.6 \\
			& UB & 1.0 & 52.3 &	76.0 &	79.6 &	63.5 &	75.2 &	76.5 &	62.1 &	49.0 &	80.7 &	69.9 &	56.4 &	83.4 &	68.7  \\
			& UB & 2.0 & 51.8 &	76.0 &	79.6 &	63.0 &	75.1 &	76.5 &	60.8 &	49.2 &	80.7 &	69.6 &	55.5 &	83.5 &	68.4 \\	
			& BR & -- &  54.2 &	\HL{79.4} &	81.5 &	66.8 &	\HL{78.6} &	79.2 &	65.6 &	50.9 &	\HL{82.6} &	71.4 &	\HL{58.1} &	85.3 &	71.1 \\	
			\midrule
			\multirow{6}{*}{kDINE$^*$} & UB & 0.0 & 54.8 &	78.4 &	81.4 &	66.8 &	77.6 &	79.2 &	67.0 &	51.6 &	82.4 &	71.8 &	58.0 &	85.2 &	71.2 \\	
			& UB & 0.1 &  54.4 &	78.0 &	80.9 &	66.5 &	76.9 &	78.9 &	66.0 &	51.1 &	82.1 &	71.4 &	57.7 &	84.7 &	70.7 \\
			& UB & 0.5 & 52.9 &	76.2 &	79.5 &	64.4 &	75.7 &	77.3 &	63.2 &	50.6 &	80.5 &	70.3 &	56.7 &	84.0 &	69.3 \\
			& UB & 1.0 & 52.1 &	75.4 &	79.2 &	63.2 &	74.9 &	75.9 &	61.6 &	48.9 &	80.3 &	69.8 &	56.0 &	83.2 &	68.4 \\
			& UB & 2.0 & 51.7 &	75.4 &	79.2 &	62.7 &	74.6 &	75.8 &	60.6 &	48.9 &	80.2 &	69.4 &	55.5 &	83.1 &	68.1 \\	
			& BR & -- &  54.2 &	78.7 &	81.4 &	66.2 &	78.1 &	78.6 &	65.0 &	50.9 &	82.3 &	71.1 &	57.5 &	85.0 &	70.7 \\	
			\bottomrule
		\end{tabular} 
	}
	\label{tab:supp:officehome_dine}
\end{table}

\section{Analysis of Prior Knowledge in kDINE}
\begin{figure}[!t]
	\centering
	\includegraphics[width=0.45\textwidth]{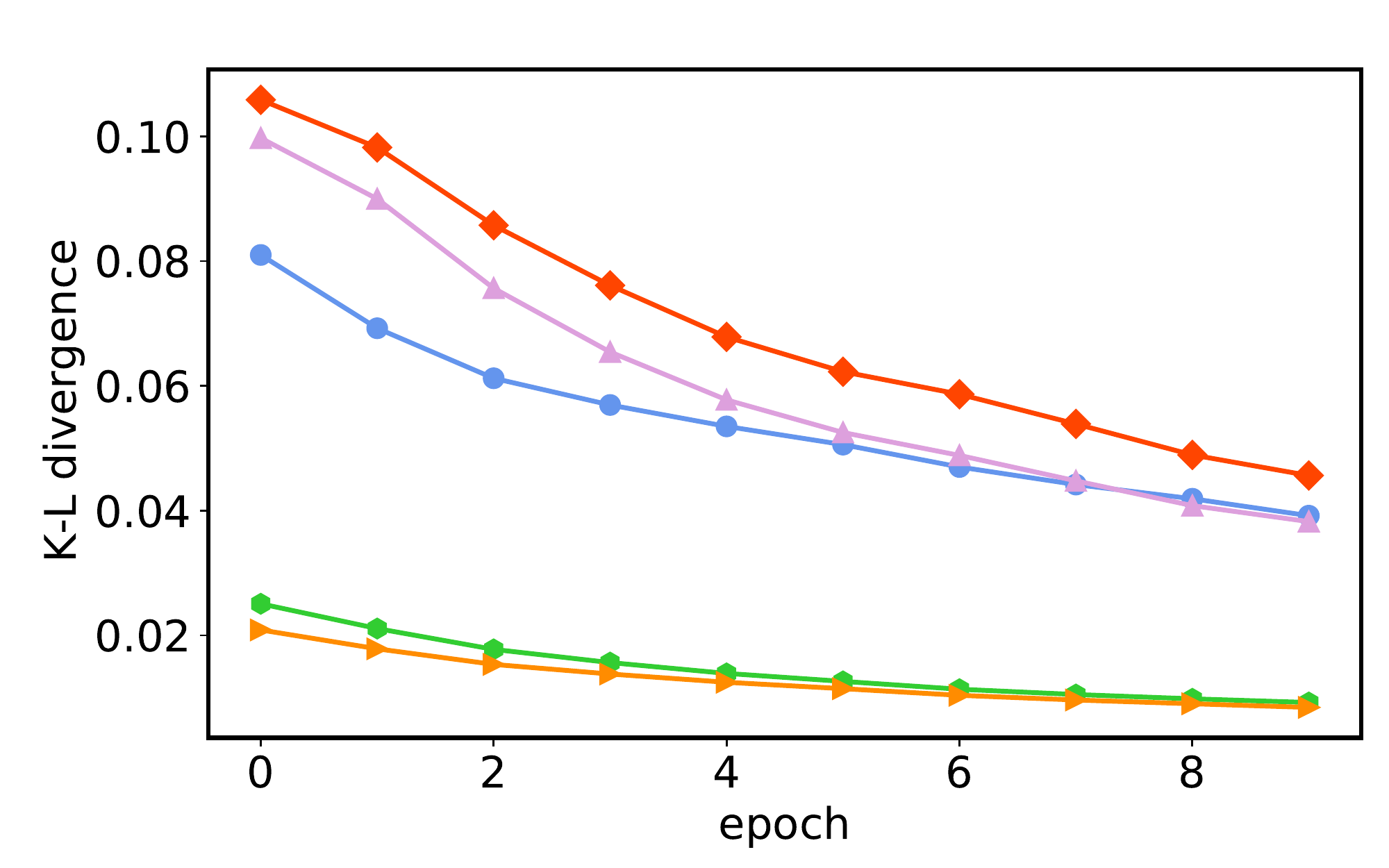}
	\includegraphics[width=0.45\textwidth]{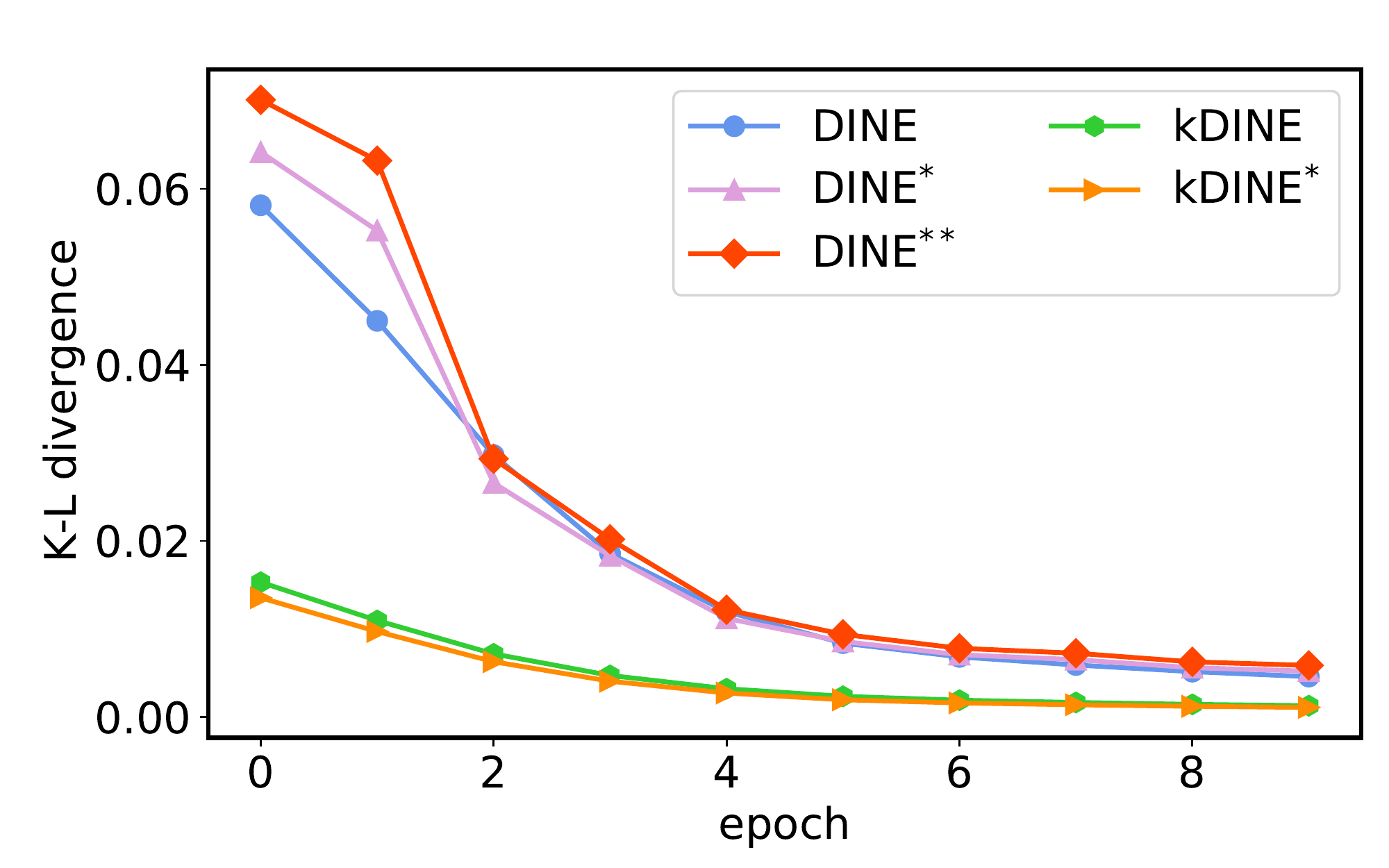}
	\caption{K-L divergences between ground-truth target class distribution and mean teacher probabilities during training on (left) Office-Home P$\rightarrow$A and (right) Office A$\rightarrow$W. (See text for details.)}
	\label{fig:supp:kl}
\end{figure}

In Fig.~\ref{fig:hist} of the paper, we show that prior knowledge rectifies distribution of pseudo labels in kSHOT. To see how prior knowledge is helpful in kDINE, let us consider the K-L divergences between the mean teacher probability and the ground-truth target class distribution. The divergences for DINE and kDINE are $\mathcal{D}_{\rm kl}\left(\mathbb{E}_{\bm{x}^t_i}\left[P^{\rm tch}(\bm{x}^t_i)\right] \Big\| p_t(y)\right)$ and $\mathcal{D}_{\rm kl}\left(\mathbb{E}_{\bm{x}^t_i}\left[\frac{P^{\rm tch}(\bm{x}^t_i)+ \tilde{\bm{l}}_i^{(\rm pk_1)}}{2}\right] \Big\| p_t(y)\right)$, respectively. Divergences for DINE$^*$, DINE$^{**}$ and kDINE$^*$ can be similarly defined. Figure~\ref{fig:supp:kl} plots these divergences on two domain adaption tasks at a step of generating teacher probability. Clearly using prior knowledge leads to much smaller divergences, which may benefit the distillation stage.

%
%
\bibliographystyle{splncs04}
\bibliography{paper}
\end{document}